\documentclass{article}

\usepackage{arxiv}

\usepackage[utf8]{inputenc} 
\usepackage[T1]{fontenc}    
\usepackage{hyperref}       
\usepackage{url}            
\usepackage{booktabs}       
\usepackage{amsfonts}       
\usepackage{nicefrac}       
\usepackage{microtype}      
\usepackage{lipsum}		
\usepackage{graphicx}
\usepackage{natbib}
\usepackage{doi}

\usepackage{bm}
\usepackage{multirow} 
\usepackage{float} 

\usepackage[most]{tcolorbox}
\usepackage{xcolor}
\usepackage{booktabs} 
\usepackage{lipsum} 
\usepackage{enumitem}
\tcbuselibrary{breakable}
\usepackage{listings}
\definecolor{navy}{HTML}{283593}
\definecolor{mygreen}{HTML}{43A047}
\definecolor{orange}{HTML}{FB8C00}
\definecolor{mydeepgreen}{RGB}{0,86,27}
\definecolor{myblue}{rgb}{0,0,1}          
\definecolor{myreddish}{rgb}{0.8,0.2,0.5} 

\usepackage{amsmath}
\usepackage{amssymb}
\usepackage{mathtools}
\usepackage{amsthm}

\title{TIDE: Tuning-Integrated Dynamic Evolution 
  for LLM-Based Automated Heuristic Design}

\date{} 					

\author{
    Chentong Chen\textsuperscript{1}\footnotemark[1] , \ 
    Mengyuan Zhong\textsuperscript{1}\thanks{Equal contribution.} , \
    Ye Fan\textsuperscript{2}, \
    Jialong Shi\textsuperscript{1}\thanks{Corresponding author.} , 
    Jianyong Sun\textsuperscript{1}\footnotemark[2] \
    \\[2mm]
    \textsuperscript{1}School of Mathematics and Statistics, Xi'an Jiaotong University, Xi'an, China \\
    \textsuperscript{2}School of Electronics and Information, Northwest Polytechnical University, Xi'an, China \\[2mm]
    \texttt{\{chengtong.chen, my.zhong\}@stu.xjtu.edu.cn, fanye@nwpu.edu.cn,} \\
    \texttt{\{jialong.shi, jy.sun\}@xjtu.edu.cn}\\
}

\begin{document}
\maketitle

\begin{abstract}
Although Large Language Models have advanced Automated Heuristic Design, treating algorithm evolution as a monolithic text generation task overlooks the coupling between discrete algorithmic structures and continuous numerical parameters. Consequently, existing methods often discard promising algorithms due to uncalibrated constants and suffer from premature convergence resulting from simple similarity metrics. To address these limitations, we propose TIDE, a Tuning-Integrated Dynamic Evolution framework designed to decouple structural reasoning from parameter optimization. TIDE features a nested architecture where an outer parallel island model utilizes Tree Similarity Edit Distance to drive structural diversity, while an inner loop integrates LLM-based logic generation with a differential mutation operator for parameter tuning. Additionally, a UCB-based scheduler dynamically prioritizes high-yield prompt strategies to optimize resource allocation. Extensive experiments across nine combinatorial optimization problems demonstrate that TIDE discovers heuristics that significantly outperform state-of-the-art baselines in solution quality while achieving improved search efficiency and reduced computational costs.
\end{abstract}


\section{Introduction}

The design of heuristic algorithms is crucial for solving NP-hard Combinatorial Optimization Problems (COPs) across various domains~\citep{sanchez2020systematic,peres2021combinatorial}. Traditionally, this process was a laborious pursuit constrained by the limits of human intuition. Recently, the advent of Large Language Models (LLMs) has catalyzed a paradigm shift, giving rise to Language Hyper-Heuristics (LHHs)~\citep{zhang2024understanding}. Groundbreaking studies, such as FunSearch~\citep{romera2024mathematical}, EoH~\citep{liu2024evolution}, and ReEvo~\citep{ye2024reevo}, have demonstrated that LLMs can act as intelligent designers~\citep{liu2024systematic} 
within evolutionary frameworks, automatically designing heuristics that surpass human-crafted baselines. By treating algorithm generation as a search problem within the code space, these methods have opened new frontiers in Automated Heuristic Design (AHD)~\citep{liu2024systematic,wu2024evolutionary}.

\begin{figure}
    \centering
    \includegraphics[width=0.6\linewidth]{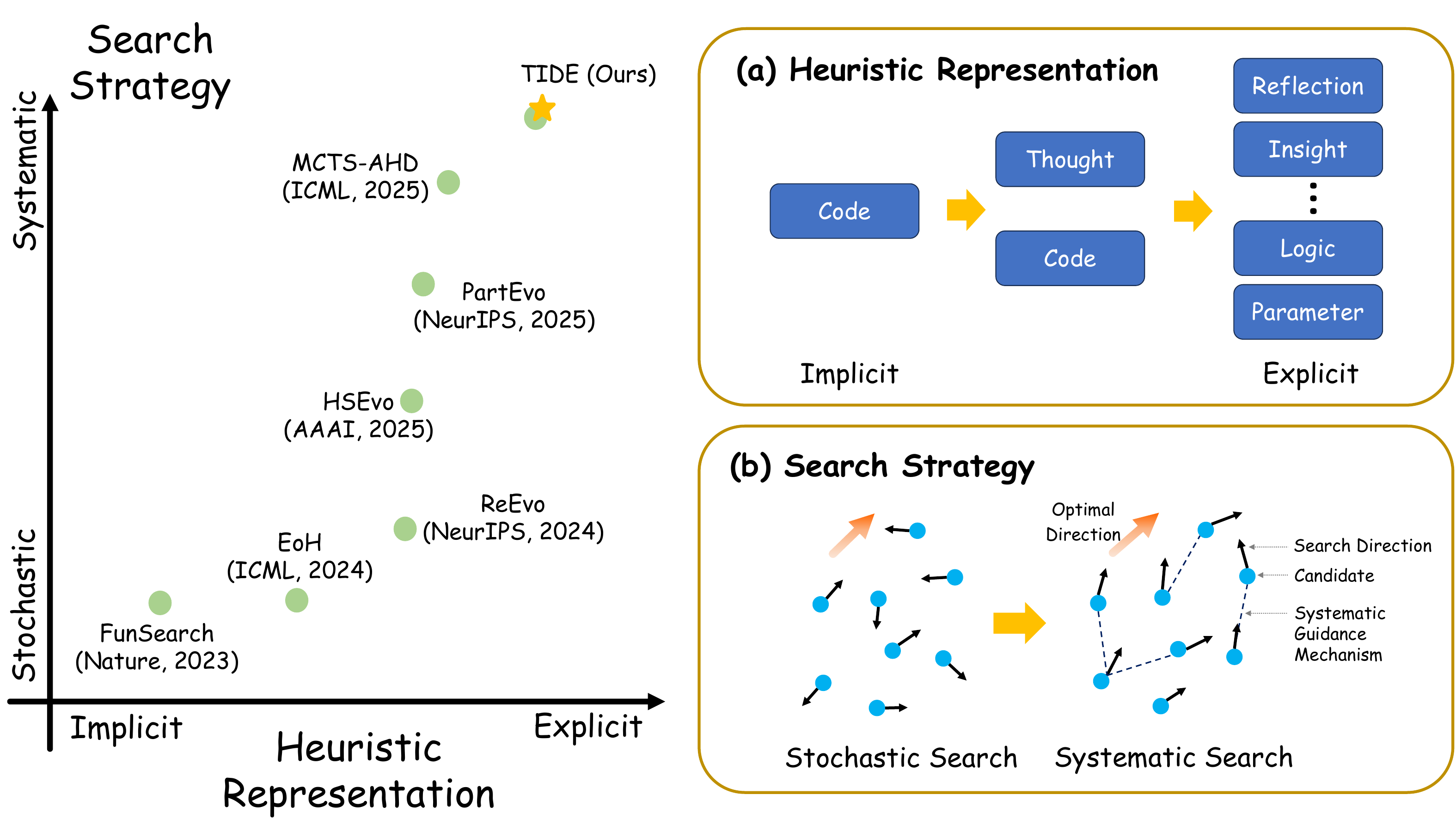}
    \caption{\textbf{Left}: Comparison of representative LLM-based AHD methods along Search Strategy (Stochastic $\to$ Systematic) and Heuristic Representation (Implicit $\to$ Explicit). \textbf{Right}: (a) \textit{Heuristic Representation}: implicit heuristics lack explicit structure, whereas explicit heuristics define modular components for interpretability and structure. (b) \textit{Search Strategy}: stochastic search uses random sampling, while systematic search employs guided mechanisms for efficient exploration.} 
    \label{fig1}
\end{figure}

However, treating algorithm evolution purely as a monolithic text generation task implies an oversimplified heuristic representation, obscuring the intrinsic nature of heuristics: the coupling of a discrete algorithmic structure (i.e., the symbolic logic flow) with a set of continuous-valued parameters. Although LLMs excel at symbolic reasoning, they suffer from \textit{numerical blindness}~\citep{feng2024numerical}, often hallucinating suboptimal constants that undermine the validity of the logic itself. The preceding approaches bifurcate the evolution of algorithmic structure and parameter configuration. For instance, AutoEP~\citep{xu2025autoep} serves as a hyperparameter controller for static algorithmic frameworks, limiting optimization to incremental refinement of pre-defined logic rather than to structural innovation. Conversely, HSEvo~\citep{dat2025hsevo} relegates parameter tuning to a post-hoc refinement stage applied only to pre-selected elites, creating a sequential dependency that overlooks the intrinsic coupling between algorithmic logic and numerical constants. Consequently, these methods bias the search towards designs that are merely robust to initialization, rather than unlocking the full potential of complex heuristics through joint optimization.

In addition, current methods struggle with both search direction control and diversity maintenance. Frameworks like EoH~\citep{liu2024evolution} and MCTS-AHD~\citep{zheng2025monte} often use random or static strategies to schedule prompts, wasting tokens on unpromising directions. Conversely, reflection-based methods~\citep{ye2024reevo,dat2025hsevo,qi2025memetic,guo2025nested} tend to converge rapidly into a single dominant logic, sacrificing population diversity. PartEvo~\citep{hu2025partition} employs niching to enhance diversity. However, it relies on textual statistics (e.g., CodeBLEU~\citep{ren2020codebleu}) or semantic embeddings~\citep{chandrasekaran2021evolution}.  These metrics cannot distinguish between real logical changes and simple code rephrasing, leading to a misaligned estimation of population diversity. 

Thus, as illustrated in Figure \ref{fig1}, many LLM-based AHD methods exhibit limitations in both heuristic representation and search strategy. To address these challenges, we propose \textbf{TIDE}, a \textbf{Tuning-Integrated Dynamic Evolution} framework that refines heuristic representation by decoupling discrete algorithmic structures from continuous numerical parameters, and optimizes search strategy through dynamic mechanisms.

TIDE adopts a nested architecture initiated by an outer \textbf{TSED-Guided Island Model}. 
To quantify algorithm structure diversity, we employ the \textit{Tree Similarity Edit Distance} (TSED)~\citep{song2024revisiting} to govern the adaptive migration and selective reset mechanisms. 
Within each island, the search is driven by a \textbf{Co-Evolutionary Inner Loop} that synergizes two specialized modules: a \textit{Upper Confidence Bound(UCB)-based} scheduler that adaptively selects high-yield prompt strategies for LLM-based logic generation, and a \textit{differential mutation operator} specifically dedicated to calibrating continuous parameters. This division of labor mitigates the LLM's numerical blindness, ensuring that evaluation fidelity reflects the true upper bound of a heuristic's performance, thereby enabling the search to accurately identify and retain high-potential algorithmic backbones.

Overall, we make the following contributions:
\begin{itemize}
\item We propose TIDE, a nested evolutionary framework that strictly decouples algorithmic logic generation from continuous parameter optimization, thereby mitigating the numerical limitations of LLMs and uncovering the performance potential of algorithmic backbones via dedicated calibration.

\item We introduce a TSED-guided Island Model that leverages a scale-invariant structural metric to govern adaptive migration and selective reset mechanisms, effectively maintaining global algorithm structure diversity and preventing premature convergence.

\item We formulate the scheduling of prompting strategies as a non-stationary Multi-Armed Bandit problem and employ a UCB-based policy to dynamically prioritize high-yield prompt strategies, maximizing search efficiency under a constrained function evaluation budget.

\end{itemize}

\section{Preliminary}

\subsection{Problem Definition: Automatic Heuristic Design}
Let $\mathcal{Q}$ denote a specific Combinatorial Optimization Problem. Associated with $\mathcal{Q}$ is an instance space $\mathcal{X}$ and a solution space $\mathcal{Y}$. We assume problem instances $x \in \mathcal{X}$ are drawn from a specific distribution $\mathcal{D}_{\mathcal{Q}}$. For any given instance $x$, a valid heuristic $h: \mathcal{X} \to \mathcal{Y}$ constructs a feasible solution $y = h(x) \in \mathcal{Y}$. The quality of this solution is evaluated by an objective function $f: \mathcal{X} \times \mathcal{Y} \to \mathbb{R}$, where $f(x, y)$ represents the cost (or negative reward) to be minimized.

The goal of AHD is to search for an optimal heuristic $h^*$ within a vast, open-ended heuristic space $\mathcal{H}$ that minimizes the expected cost over the problem distribution~\citep{ye2024reevo,zheng2025monte}:
\begin{equation}
    h^* = \mathop{\arg\min}_{h \in \mathcal{H}} \mathbb{E}_{x \sim \mathcal{D}_{\mathcal{Q}}} \left[ f(x, h(x)) \right].
\end{equation}
In practice, since the true distribution $\mathcal{D}_{\mathcal{Q}}$ is typically unknown or continuous, the expectation is approximated by the empirical average performance on a finite training set of instances $\{x_1, \dots, x_N\}$ sampled from $\mathcal{D}_{\mathcal{Q}}$.

\subsection{Program Representation and Structural Distance}
\label{sec:representation_metric}

Unlike neural policies operating in latent spaces, heuristics in AHD are symbolic programs where functionality is coupled with syntax. To robustly quantify phenotypic diversity, we utilize a structural analysis pipeline that decouples algorithmic logic from surface-level lexical variations.

\paragraph{Abstract Syntax Tree (AST) and Tree Edit Distance (TED).}
Let $\mathcal{C}$ denote the space of source code. We define a parsing and normalization function $\Phi: \mathcal{C} \to \mathbb{T}$ that maps a heuristic $c$ to an Abstract Syntax Tree (AST) $\mathcal{T} = (V, E, \lambda)$. Here, $V$ is the set of nodes, $E$ represents hierarchical parent-child dependencies, and $\lambda: V \to \Sigma$ assigns labels from a simplified vocabulary $\Sigma$.
To measure the topological divergence between two heuristics, we employ Tree Edit Distance (TED)~\citep{zhang1989simple}, computed via the state-of-the-art APTED algorithm~\citep{pawlik2015efficient}. TED is defined as the minimum cost of a sequence of elementary operations, comprising insertion, deletion, and renaming, that transforms a source tree $\mathcal{T}_a$ into a target tree $\mathcal{T}_b$.
Formally, let $\mathcal{S}_{a \to b}$ be the set of all valid edit scripts transforming $\mathcal{T}_a$ to $\mathcal{T}_b$. The edit distance $\Delta$ is the minimization objective:
\begin{equation}
    \Delta(\mathcal{T}_a, \mathcal{T}_b) = \min_{S \in \mathcal{S}_{a \to b}} \sum_{op \in S} \gamma(op),
\end{equation}
where $\gamma(op)$ represents the cost function associated with operation $op$ (typically unit cost). 

\section{TIDE-AHD}

\begin{figure*}
    \centering
    \includegraphics[width=\linewidth]{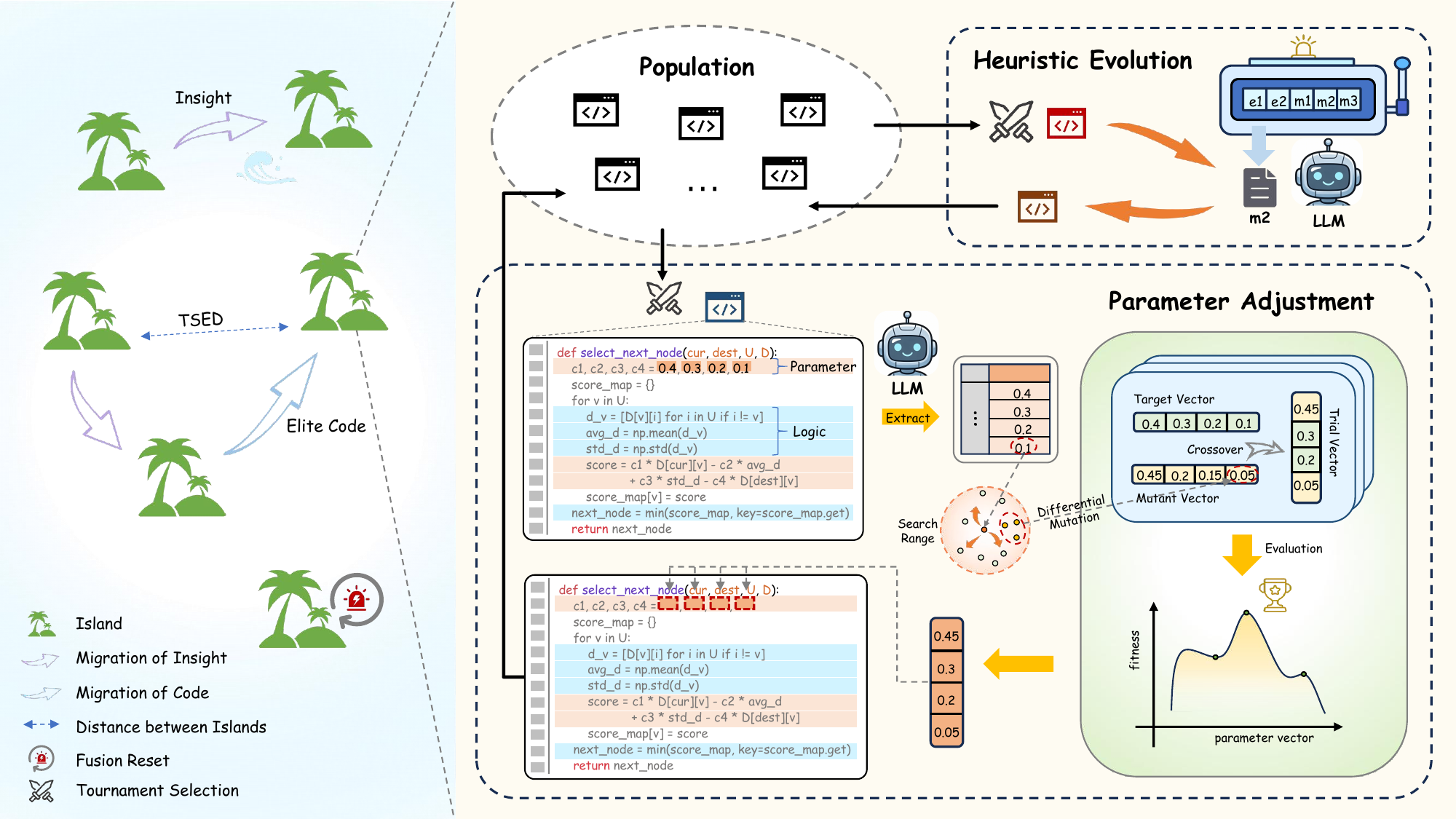}
    \caption{The pipeline of TIDE-AHD. \textbf{Left:}The outer loop functions as a TSED-guided Island Model, regulating global diversity via adaptive migration and fusion resets.  \textbf{Right:} The inner loop executes co-evolutionary search, synergizing UCB-scheduled algorithmic logic generation with parameter refinement to mitigate numerical blindness.}
    \label{fig:pipeline}
\end{figure*}

We propose \textbf{TIDE}, a framework instantiated as a nested evolutionary architecture (Figure~\ref{fig:pipeline}). As implied by the nomenclature, the system orchestrates two distinct levels of optimization: the \textit{Dynamic Evolution} corresponds to an \textit{Outer Loop} that navigates the global topological landscape through a TSED-guided Island Model, while the \textit{Tuning-Integrated} component refers to an \textit{Inner Loop} that synergizes algorithmic structure reasoning with precise parameter calibration. 
\subsection{TSED-Guided Island Model}
\label{sec:outer}
\subsubsection{Structural Similarity Quantification using TSED}

Current AHD methods predominantly follow a \textit{Thought-to-Code} paradigm, often approximating diversity via the textual similarity of intermediate thoughts or superficial lexical metrics like Jaccard distance. We contend that these proxies are ill-suited for heuristic evolution, as they fail to distinguish between lexical variations and genuine algorithmic novelty. To quantify phenotypic diversity, we employ the \textit{Tree Similarity Edit Distance} (TSED)~\citep{song2024revisiting}, a metric that operates on the AST level to isolate control flow from syntactic noise. 

The computation of TSED follows a deterministic pipeline designed to ensure evaluation fidelity. First, the generated heuristic code $c$ is parsed into a raw AST. To mitigate the stochasticity of LLM-generated identifiers, we apply a structural normalization function $\Phi(\cdot)$ that removes non-functional elements such as comments and imports, and maps specific variable names to uniform placeholders. This yields a normalized tree $\hat{\mathcal{T}} = \Phi(c)$ representing the algorithmic backbone. Subsequently, we employ the APTED algorithm to compute the TED, denoted as \textbf{$\Delta(\hat{\mathcal{T}}_A, \hat{\mathcal{T}}_B)$}, which quantifies the minimum cumulative cost of elementary operations, including insertion, deletion, and renaming, required to align two normalized trees. Finally, to ensure scale invariance across heuristics of varying complexity, we derive the normalized similarity score $S_{\text{TSED}} \in [0, 1]$ as:
\begin{equation}
S_{\text{TSED}}(\hat{\mathcal{T}}_A, \hat{\mathcal{T}}_B) = \max \left( 0, 1 - \frac{\Delta(\hat{\mathcal{T}}_A, \hat{\mathcal{T}}_B)}{\max(|\hat{\mathcal{T}}_A|, |\hat{\mathcal{T}}_B|)} \right),
\end{equation}
where $|\hat{\mathcal{T}}|$ denotes the node count of the normalized tree. This formulation effectively filters out superficial textual differences, providing a robust basis for the island model's migration decisions.

\subsubsection{Adaptive Migration and Reset Strategies}
\label{sec:migration}

\textbf{TSED-based Topology Construction.}
In our Island Model, TSED serves as the definitive metric for defining the interaction topology among subpopulations. Let $\mathcal{I} = \{I_1, I_2, \dots, I_K\}$ denote the set of $K$ independent islands. Within each island $I_k$, the population consists of a set of individuals, where each individual is defined as a heuristic algorithm represented by Python code. At each migration interval, we construct a pairwise code similarity matrix $\mathbf{M} \in [0,1]^{K \times K}$ to quantify structural relationships across the global population. The entry $M_{ij}$ is computed as the arithmetic mean of the TSED similarity scores between the individuals of island $i$ and island $j$:
\begin{equation}
M_{ij} = \frac{1}{|I_i||I_j|} \sum_{c \in I_i} \sum_{c' \in I_j} S_{\text{TSED}}(c, c')
\end{equation}

This matrix provides a comprehensive topological map of the search space, distinguishing between islands that have collectively converged into similar algorithmic basins and those that explore distinct structural regions.

\textbf{Dual-Mode Migration Strategy.}
Migration is triggered adaptively to counteract stagnation, utilizing a dual-mode transfer strategy governed by the structural topology matrix $\mathbf{M}$. For island pairs exhibiting high structural convergence (high $M_{ij}$), which signals a shared basin of attraction, maintaining independence becomes computationally redundant. In such cases, we enforce exploitation by directly overwriting the target island's lowest-performing individual with the source island's current best-performing individual (the elite). This synchronizes fine-grained parameter refinements across structurally similar populations. Conversely, for pairs located in distinct structural basins (low $M_{ij}$), direct code transfer risks disrupting local evolutionary trajectories. Therefore, we prioritize exploration through conceptual knowledge transfer: the LLM distills the pivotal algorithmic mechanisms of the source elite into a natural language insight~\citep{chen2025hifo}, injecting it into the target's context to assimilate superior design principles without compromising its unique implementation architecture.

\textbf{Constructive Fusion Reset.}
Reset serves as a rigorous intervention reserved for deep stagnation where an island fails to improve over an extended timeframe. Rather than random re-initialization, we employ a \textit{Constructive Fusion Mechanism}. We prompt the LLM to synthesize a hybrid candidate by integrating the robust structural backbone of the current global elite with the specialized local mechanisms of the Stagnated Incumbent. This strategy generates a high-quality seed that preserves accumulated domain knowledge while injecting sufficient structural novelty to re-energize the search.

\subsection{Co-Evolution of Algorithm Structure and Parameters}
\label{sec:inner}



We formalize the heuristic space as $\mathcal{H} = \bigcup_{l \in \mathcal{L}} \{ \Phi(l,p) \mid p \in \mathcal{P}(l) \subseteq \mathbb{R}^{d(l)} \}$, where each heuristic combines a discrete algorithmic structure $l$(encoding the symbolic logic) with continuous parameters $p$. Consequently, the AHD objective decomposes into a bi-level optimization problem:
\begin{equation}
    h^* = \operatorname*{argmin}_{l \in \mathcal{L}} \left( \min_{p \in \mathcal{P}(l)} \mathbb{E}_{x \sim \mathcal{D}_{\mathcal{Q}}} [f(x, \Phi(l,p)(x))] \right).
\end{equation}
This structure motivates our framework's division into two complementary components: \emph{Structural Search} to traverse the logic space $\mathcal{L}$, and \emph{Parameter Refinement} to optimize $p$ within the selected logic.

\subsubsection{Structural Search with Adaptive Prompting Strategy}

\textbf{Prompting Action Space.}
In LLM-based AHD, the structural quality of generated offspring is intrinsically linked to the specific prompting strategy employed. We formally define a discrete action space $\mathcal{A} = \{ a_1, a_2, \dots, a_K \}$, where each action $a_k$ constitutes a distinct prompting strategy, serving as an evolutionary operator instantiated through a specific prompt template to elicit targeted algorithmic transformations. 
Drawing upon the paradigms established in EoH \cite{liu2024evolution}, our action space is categorized into crossover and mutation strategies. The crossover strategies include \textit{Exploratory Generation} ($e_1$), which synthesizes novel architectures to maximize diversity, and \textit{Backbone Consensus} ($e_2$), designed to extract and preserve shared high-performing substructures. The mutation strategies consist of \textit{Topology Perturbation} ($m_1$) for modifying control flows, \textit{Scoring Logic Refinement} ($m_2$) for adjusting internal weighting schemes, and \textit{Parsimony Enforcement} ($m_3$) for simplifying logic to enhance generalization capabilities.

\textbf{Stochastic MAB Formulation.}
Selecting the optimal strategy is non-trivial due to two intrinsic properties of the generative process. First, LLM-based generation is stochastic: invoking a strategy $a_k$ yields samples from a high-variance distribution rather than deterministic improvements, resulting in noisy reward signals. Second, the process is non-stationary: the marginal utility of a specific strategy is not constant but evolves dynamically depending on the optimization state.

Consequently, we formulate this sequential decision-making process as a \textit{Multi-Armed Bandit (MAB)} problem~\citep{chen2013combinatorial,besbes2014stochastic}. In this setting, the prompting strategies in $\mathcal{A}$ correspond to the arms, and the normalized fitness improvement serves as a stochastic reward signal. The central challenge constitutes an exploration-exploitation dilemma: given a constrained budget of function evaluations, the search process must strategically balance the exploitation of strategies with proven efficacy against the exploration of under-sampled prompt strategies.

\textbf{UCB Selection Policy.}
We address this dilemma by implementing the \textit{Upper Confidence Bound (UCB)} algorithm~\citep{auer2002finite,dacosta2008adaptive,candan2013non,barto2021reinforcement}. This deterministic policy selects the prompting strategy $a_t$ at iteration $t$ by maximizing a score that accounts for both the empirical reward estimate and the uncertainty of that estimate:
\begin{equation}
a_t = \operatorname*{argmax}_{a \in \mathcal{A}} \left( Q_t(a) + C \sqrt{\frac{2 \ln N_t}{n_t(a)}} \right)
\end{equation}

where $Q_t(a)$ denotes the average reward obtained by strategy $a$ up to iteration $t$, quantified as the relative fitness gain over the current optimum; $N_t$ and $n_t(a)$ represent the global count of strategy invocations and the specific selection frequency of strategy $a$, respectively; and $C$ is the exploration constant that modulates the confidence interval width to control the propensity for testing uncertain strategies. By dynamically updating these statistics, UCB effectively mitigates the impact of noisy rewards and adapts the prompt strategies distribution to the evolving optimization landscape.

\textbf{Evaluation and Update.}
For crossover strategies, we utilize \textit{Tournament Selection} to sample parents from the local subpopulation. Conversely, mutation strategies prioritize the incumbent elite to leverage the existing algorithmic backbone. Following code generation and evaluation, the valid offspring are merged into the subpopulation. We then enforce a rank-based survival strategy where the pool is sorted and truncated to retain only the top-performing individuals. The cycle concludes by calculating the relative fitness gain of the offspring, which serves as the reward signal to update the value estimates of the UCB policy.

\subsubsection{Parameter Refinement with Differential Mutation Operator}

Effective heuristic algorithms rely on the synergy between discrete symbolic logic and continuous hyperparameters, where performance is often critically sensitive to tightly coupled numerical values. Although LLMs excel at semantic reasoning for structural discovery, their nature as probabilistic token predictors limits their arithmetic precision, rendering them inefficient for iterative numerical fine-tuning. To resolve this, we draw upon the design philosophy of \textit{Memetic Algorithms (MA)}~\citep{krasnogor2005tutorial,neri2011handbook}, which advocates coupling global structural search with intensive local refinement.  

Specifically, we employ a \textbf{Differential Mutation Operator} adapted from standard Differential Evolution (DE)\citep{storn1997differential}. Given the gradient-free nature of the heuristic optimization landscape, this operator effectively steers the search direction using scaled vector differences between population members. This mechanism provides an adaptive means of navigating the complex terrain of coupled parameters, enabling efficient fine-tuning of the novel heuristic logic proposed by the LLM. 

\begin{table*}[t]
\centering
\caption{
Performance of constructive and improvement heuristics on multiple combinatorial optimization tasks.
Obj.$\downarrow$ / $\uparrow$ indicate minimization / maximization objectives, and Gap reports deviation from the corresponding optimal or best-known reference.
Optimal results are obtained using LKH3 (TSP) and OR-Tools (KP), and optimal results for online BPP and ASP are taken from MCTS-AHD. All results of LLM-based methods are averaged over three runs. The best results among LLM-based AHD methods are highlighted in bold.
}

  \resizebox{\textwidth}{!}{
    \begin{tabular}{l|rrrrrr|rrrrrr}
    \toprule
    \multicolumn{13}{c}{Constructive Heuristic} \\
    \midrule
    \multicolumn{1}{c|}{Task} & \multicolumn{6}{c|}{TSP}                      & \multicolumn{6}{c}{KP} \\
    \midrule
    \multicolumn{1}{c|}{Scale} & \multicolumn{2}{c}{N=50} & \multicolumn{2}{c}{N=100} & \multicolumn{2}{c|}{N=200} & \multicolumn{2}{c}{N=50, W=12.5} & \multicolumn{2}{c}{N=100, W=25} & \multicolumn{2}{c}{N=200, W=25} \\
    \multicolumn{1}{c|}{Methods} & \multicolumn{1}{c}{Obj.↓} & \multicolumn{1}{c}{Gap} & \multicolumn{1}{c}{Obj.↓} & \multicolumn{1}{c}{Gap} & \multicolumn{1}{c}{Obj.↓} & \multicolumn{1}{c|}{Gap} & \multicolumn{1}{c}{Obj.↑} & \multicolumn{1}{c}{Gap} & \multicolumn{1}{c}{Obj.↑} & \multicolumn{1}{c}{Gap} & \multicolumn{1}{c}{Obj.↑} & \multicolumn{1}{c}{Gap} \\
    \midrule
    Optimal & 5.68  & 0.00\% & 7.75  & 0.00\% & 10.72 & 0.00\% & 20.04 & 0.00\% & 40.27 & 0.00\% & 57.45 & 0.00\% \\
    Greedy & 6.96  & 22.62\% & 9.71  & 25.32\% & 13.46 & 25.60\% & 19.99 & 0.26\% & 40.23 & 0.12\% & 57.40 & 0.09\% \\
    POMO  & 5.70  & 0.38\% & 8.00  & 3.30\% & 12.90 & 20.33\% & 19.61 & 2.12\% & 39.68 & 1.48\% & 57.27 & 0.31\% \\
    \midrule
    EoH   & 6.36  & 12.11\% & 8.89  & 14.72\% & 12.45 & 16.14\% & 19.92 & 0.60\% & 40.05 & 0.55\% & 56.44 & 1.76\% \\
    ReEvo & 6.15  & 8.27\% & 8.52  & 10.06\% & 11.87 & 10.75\% & 20.00 & 0.21\% & 40.23 & 0.10\% & 57.40 & 0.08\% \\
    HSEvo & 6.13  & 8.04\% & 8.52  & 10.01\% & 11.88 & 10.80\% & 20.00 & 0.17\% & 40.24 & 0.08\% & 57.41 & 0.07\% \\
    MCTS-AHD & 6.14  & 8.19\% & 8.53  & 10.18\% & 11.91 & 11.09\% & 20.00 & 0.17\% & 40.24 & 0.08\% & 57.41 & 0.07\% \\
    TIDE(ours) & \textbf{5.95} & \textbf{4.76\%} & \textbf{8.30} & \textbf{7.15\%} & \textbf{11.86} & \textbf{10.65\%} & \textbf{20.03} & \textbf{0.04\%} & \textbf{40.26} & \textbf{0.02\%} & \textbf{57.44} & \textbf{0.02\%} \\
    \midrule
    \multicolumn{1}{c|}{Task} & \multicolumn{6}{c|}{BPP online}               & \multicolumn{6}{c}{ASP} \\
    \midrule
    \multicolumn{1}{c|}{Scale} & \multicolumn{2}{c}{N=1k, W=100} & \multicolumn{2}{c}{N=5k, W=100} & \multicolumn{2}{c|}{N=10k, W=100} & \multicolumn{2}{c}{n=12, w=7} & \multicolumn{2}{c}{n=15, w=10} & \multicolumn{2}{c}{n=21, w=15} \\
    \multicolumn{1}{c|}{Methods} & \multicolumn{1}{c}{Obj.↓} & \multicolumn{1}{c}{Gap} & \multicolumn{1}{c}{Obj.↓} & \multicolumn{1}{c}{Gap} & \multicolumn{1}{c}{Obj.↓} & \multicolumn{1}{c|}{Gap} & \multicolumn{1}{c}{Obj.↑} & \multicolumn{1}{c}{Gap} & \multicolumn{1}{c}{Obj.↑} & \multicolumn{1}{c}{Gap} & \multicolumn{1}{c}{Obj.↑} & \multicolumn{1}{c}{Gap} \\
    \midrule
    Optimal & 402.4 & 0.00\% & 2019.4 & 0.00\% & 4010.6 & 0.00\% & 792   & 0.00\% & 3003  & 0.00\% & 43596 & 0.00\% \\
    \midrule
    EoH   & 418.2 & 3.93\% & 2043.6 & 1.20\% & 4035.9 & 0.63\% & 775   & 2.15\% & 2732  & 9.02\% & 32217 & 26.10\% \\
    ReEvo & 418.5 & 4.01\% & 2088.1 & 3.40\% & 4139.9 & 3.22\% & 767   & 3.16\% & 2781  & 7.39\% & 29743 & 31.78\% \\
    HSEvo & 421.5 & 4.74\% & 2104.7 & 4.22\% & 4172.1 & 4.03\% & 760   & 4.04\% & 2747  & 8.52\% & 30291 & 30.52\% \\
    MCTS-AHD & 417.5 & 3.76\% & 2069.7 & 2.49\% & 4095.9 & 2.13\% & 777   & 1.89\% & 2801  & 6.73\% & 33334 & 23.54\% \\
    TIDE(ours) & \textbf{415.8} & \textbf{3.33\%} & \textbf{2034.2} & \textbf{0.73\%} & \textbf{4025.1} & \textbf{0.36\%} & \textbf{780} & \textbf{1.52\%} & \textbf{2812} & \textbf{6.36\%} & \textbf{33549} & \textbf{23.04\%} \\
    \midrule
    \multicolumn{13}{c}{Improvement Heuristic} \\
    \midrule
    \multicolumn{1}{c|}{Task} & \multicolumn{6}{c|}{TSP GLS}                  & \multicolumn{6}{c}{TSP KGLS} \\
    \midrule
    \multicolumn{1}{c|}{Scale} & \multicolumn{2}{c}{N=100} & \multicolumn{2}{c}{N=200} & \multicolumn{2}{c|}{N=500} & \multicolumn{2}{c}{N=100} & \multicolumn{2}{c}{N=200} & \multicolumn{2}{c}{N=500} \\
    \multicolumn{1}{c|}{Methods} & \multicolumn{1}{c}{Obj.↓} & \multicolumn{1}{c}{Gap} & \multicolumn{1}{c}{Obj.↓} & \multicolumn{1}{c}{Gap} & \multicolumn{1}{c}{Obj.↓} & \multicolumn{1}{c|}{Gap} & \multicolumn{1}{c}{Obj.↓} & \multicolumn{1}{c}{Gap} & \multicolumn{1}{c}{Obj.↓} & \multicolumn{1}{c}{Gap} & \multicolumn{1}{c}{Obj.↓} & \multicolumn{1}{c}{Gap} \\
    \midrule
    Optimal & 7.76  & 0.000\% & 10.70 & 0.00\% & 16.52 & 0.00\% & 7.768  & 0.000\% & 10.71  & 0.00\% & 16.50  & 0.00\% \\
    NeuOpt & 7.78  & 0.304\% & 10.82 & 1.08\% & \multicolumn{1}{c}{-} & \multicolumn{1}{c|}{-} & 7.796  & 0.362\% & 10.82  & 1.05\% & \multicolumn{1}{c}{-} & \multicolumn{1}{c}{-} \\
    \midrule
    EoH   & 7.76  & 0.034\% & 10.74 & 0.37\% & 16.75 & 1.37\% & \textbf{7.768 } & \textbf{0.003\%} & 10.73  & 0.23\% & 16.66  & 0.96\% \\
    ReEvo & 7.82  & 0.862\% & 10.98 & 2.56\% & 17.14 & 3.74\% & 7.769  & 0.010\% & 10.75  & 0.36\% & 16.72  & 1.36\% \\
    HSEvo & 7.76  & 0.064\% & 10.79 & 0.76\% & 16.86 & 2.04\% & 7.769  & 0.008\% & 10.73  & 0.22\% & 16.66  & 0.97\% \\
    MCTS-AHD & 7.76  & 0.081\% & 10.76 & 0.49\% & 16.79 & 1.62\% & 7.769  & 0.011\% & 10.73  & 0.23\% & 16.66  & 0.98\% \\
    TIDE(ours) & \textbf{7.76} & \textbf{0.026\%} & \textbf{10.74} & \textbf{0.34\%} & \textbf{16.75} & \textbf{1.37\%} & 7.768  & 0.006\% & \textbf{10.73 } & \textbf{0.20\%} & \textbf{16.65 } & \textbf{0.90\%} \\
    \bottomrule
    \end{tabular}%
    }
  \label{tab:1}%
\end{table*}%

\begin{table*}[t]
  \centering
\caption{
Performance of different heuristics with different frameworks on various tasks. The ACO framework is evaluated on TSP, CVRP, OP, MKP, and offline BPP; the GA framework on DPP; RL methods on Mountain Car. The results in the table represent objective function values, and
arrows ($\downarrow$ / $\uparrow$) indicate minimization/maximization objectives. The results of ACO and DeepACO are taken from MCTS-AHD and marked with an asterisk (*). LLM-based results are averaged over three runs. And, the best results are highlighted in bold.
}
  \resizebox{\textwidth}{!}{
    \begin{tabular}{l|rr|rr|rr|rr|rr|c|c}
    \toprule
    \multicolumn{1}{c|}{Frameworks} & \multicolumn{10}{c|}{ACO}                                                     & GA    & Others \\
    \midrule
    \multicolumn{1}{c|}{Task} & \multicolumn{2}{c|}{TSP↓} & \multicolumn{2}{c|}{CVRP↓} & \multicolumn{2}{c|}{OP↑} & \multicolumn{2}{c|}{MKP↑} & \multicolumn{2}{c|}{BPP offline↓} & DPP↑  & Mountain Car↓ \\
    \multicolumn{1}{c|}{Scale} & \multicolumn{1}{c}{50} & \multicolumn{1}{c|}{100} & \multicolumn{1}{c}{50} & \multicolumn{1}{c|}{100} & \multicolumn{1}{c}{100} & \multicolumn{1}{c|}{200} & \multicolumn{1}{c}{100} & \multicolumn{1}{c|}{200} & \multicolumn{1}{c}{500} & \multicolumn{1}{c|}{1000} & -     & - \\
    \midrule
    ACO   & 5.99  & 8.95  & 11.36 & 18.78 & 24.11 & 36.83 & 22.74 & 40.67 & 208.83 & 417.94 & -     & - \\
    DeepACO & 5.84  & 8.28  & 8.89  & 14.93 & 30.36 & 51.35 & 23.09 & 41.99 & 203.13 & 405.17 & -     & - \\
    EoH   & 5.85  & 8.35  & 9.25  & 15.96 & 29.94 & 53.17 & 23.22 & 42.30 & 204.59 & 408.48 & -     & 103.3 \\
    ReEvo & 5.84  & 8.32  & 9.80  & 16.68 & 30.05 & 52.79 & 23.23 & 42.36 & 204.70 & 408.73 & 12.74 & 106.3 \\
    HSEvo & 5.84  & 8.28  & 9.43  & 16.01 & 30.79 & 54.70 & 23.26 & 42.45 & 204.50 & 408.36 & -     & 116.7 \\
    MCTS-AHD & 5.82  & 8.20  & 9.12  & 15.85 & 30.73 & 54.74 & 23.30 & 42.77 & 203.36 & 405.63 & -     & 105.7 \\
    TIDE(ours) & \textbf{5.80} & \textbf{8.15} & \textbf{9.01} & \textbf{15.37} & \textbf{30.90} & \textbf{55.43} & \textbf{23.33} & \textbf{42.90} & \textbf{203.20} & \textbf{405.20} & \textbf{12.80} & \textbf{98.3} \\
    \bottomrule
    \end{tabular}%
    }
  \label{tab:2}%
\end{table*}%

\textbf{Conditional Activation and Trust Region Construction.}
We restrict intensive local refinement to offspring competitive with the island's current elite. This protocol filters out structurally inferior candidates early, concentrating gradient-free optimization resources solely on high-potential algorithm structures.

Once activated, we employ the LLM to identify tunable parameters in the code and instantiate the initial parameter vector $\mathbf{x}_{\text{init}}$. 
To ensure subsequent variations remain semantically valid, we construct a \textit{dynamic hyper-rectangular trust region} $\Omega$ centered on $\mathbf{x}_{\text{init}}$. Since heuristic parameters often vary across multiple orders of magnitude, we adaptively define the search boundaries based on the initial value of each dimension. For standard coefficients, we determine the upper and lower bounds proportional to the parameter's magnitude, allowing for sufficient directional exploration without destabilizing the logic. In contrast, for near-zero parameters where proportional bounds would become negligible, we apply a constant value range to maintain an effective search volume. Finally, to bootstrap the local search, we initialize a micro-population $\mathcal{P}$ by applying \textit{Gaussian perturbations}~\citep{sun2019differential} to $\mathbf{x}_{\text{init}}$, thereby exploring the local neighborhood while preserving the original algorithmic intent.

\textbf{Differential Mutation Mechanism.}
Operating on this initialized micro-population, we execute the canonical \texttt{rand/1} differential mutation scheme~\citep{storn1997differential}. For a target vector $\mathbf{x}_i$, a mutant vector $\mathbf{v}_i$ is generated as :
\begin{equation} \label{eq:mutation}
\mathbf{v}_i = \mathbf{x}_{r1} + F \cdot (\mathbf{x}_{r2} - \mathbf{x}_{r3})
\end{equation}
where $\mathbf{x}_{r1}, \mathbf{x}_{r2}, \mathbf{x}_{r3}$ are three distinct vectors sampled uniformly from $\mathcal{P}$. The difference vector $(\mathbf{x}_{r2} - \mathbf{x}_{r3})$ establishes a stochastic search direction derived from the local neighborhood geometry, while the scaling factor $F \in (0, 1)$ moderates the mutation strength.

We then apply binomial crossover to produce the final trial vector $\mathbf{u}_i$:
\begin{equation} \label{eq:crossover}
u_{i,j} = 
\begin{cases} 
v_{i,j} & \text{if~~} \mathcal{R}_j \le \text{\textit{CR}~~or~~} j = j_{\text{rand}} \\
x_{i,j} & \text{otherwise}
\end{cases}
\end{equation}
where, for each dimension $j$, $\mathcal{R}_j \sim \mathcal{U}(0, 1)$. The crossover rate \textit{CR} $\in [0, 1]$ controls the probability of inheriting components from the mutant vector, ensuring that at least one dimension ($j_{\text{rand}}$) is updated to drive the search forward.

\textbf{Evaluation and Update.}
We generate and evaluate a candidate batch in parallel by mapping trial vectors back into the executable structure to obtain their objective costs $g(\cdot)$. Adopting a greedy selection strategy, we update the current parameter vector $\mathbf{x}_i$ with the optimal candidate $\mathbf{u}^* = \operatorname*{argmin}_{k} g(\mathbf{u}^{(k)})$ only if it improves the solution quality (i.e., $g(\mathbf{u}^*) < g(\mathbf{x}_i)$). This update mechanism ensures that novel algorithmic architectures are paired with optimized numerical configurations independent of LLM inference. Furthermore, these optimized constants are preserved in the global elite population, propagating fine-grained numerical traits to accelerate convergence across the outer Island Model.

\section{Experiments}

In this section, we evaluate the heuristics designed by our method across 9 problem types and 5 algorithmic frameworks. The problem types include several well-known NP-hard combinatorial optimization problems and more challenging complex optimization problems.
The problem formulations are detailed in Appendix \ref{app:problem}. We consider 5 general algorithmic frameworks in our experiments, including Constructive Heuristic, Improvement Heuristic, Ant Colony Optimization (ACO), Genetic Algorithm (GA), and Reinforcement Learning (RL). These frameworks are introduced in Appendix \ref{app:frameworks}. 

\subsection{Experimental Setup}

\textbf{Setting.} For fair comparison, all experiments were conducted on an Intel Core i7-12700 CPU, employing \texttt{Qwen3-Max-2025-09-23} accessed via its official API as the unified LLM for all methods. The outer framework employs 6 islands with a TSED diversity threshold of 0.7. In the inner framework, each island has a population size of 8 and runs for 800 generations. For the adaptive operator selection strategy, the UCB exploration constant is set to $\sqrt{2}$, following UCB1~\citep{auer2002finite}. A more detailed description is provided in Appendix \ref{app:setting}.

\textbf{Baseline.}
To provide an intuitive comparison of the performance of heuristics designed by our method, we introduce several methods from different domains as baselines. (1) Handcrafted heuristics, e.g., LKH3~\citep{lin1973effective} for TSP, OR-Tools for KP. (2) Neural Combinatorial Optimization (NCO) models, e.g., POMO~\citep{kwon2020pomo} for TSP and KP in a constructive heuristic framework, NeuOpt~\citep{ma2023learning} for TSP in an improvement heuristic framework, DeepACO~\citep{ye2023deepaco} for TSP, CVRP, OP, MKP and BPP offline. (3) LLM-based AHD methods, e.g., EoH~\citep{liu2024evolution}, ReEvo~\citep{ye2024reevo}, HSEvo~\citep{dat2025hsevo} and MCTS-AHD~\citep{zheng2025monte}.

\subsection{Empirical Result}

\paragraph{Constructive Heuristic.}
Constructive heuristics generate solutions incrementally by making a sequence of decisions, where each decision determines how a partial solution is extended. In this framework, we evaluate heuristics on TSP, KP, online BPP, and ASP. Our method is to automatically design construction rules that guide the selection of the next element to be added to the partial solution. Table~\ref{tab:1} reports the performance of different heuristic design methods in constructive frameworks. Overall, the heuristic designed by our method consistently achieves the best performance in LLM-based methods across all evaluated tasks and problem scales. Notably, on the KP task, our automatically designed heuristics outperform end-to-end NCO models.



\paragraph{Improvement Heuristic.}

Improvement heuristics start from an initial feasible solution and iteratively refine it through local search operations. Among them, Guided Local Search (GLS) and its variants, like Knowledge-Guided Local Search (KGLS), are widely adopted for large-scale combinatorial optimization due to their strong empirical performance and clear separation between search operators and guidance mechanisms. We consider TSP under the GLS and KGLS frameworks. Table~\ref{tab:1} reports the results on TSP under both the GLS and KGLS frameworks. The heuristics designed by our method exhibit competitive performance, outperforming heuristics generated by existing LLM-based AHD methods and achieving substantial performance advantages over the improvement-based NCO model.

\paragraph{Ant Colony Optimization.}

Ant Colony Optimization (ACO) is a population-based search framework that generates solutions by simulating the collective behavior of ants guided by pheromone trails and heuristic information. This setup allows us to assess the generality of our method across multiple problem domains under a unified search framework. So, we evaluate our method on TSP, CVRP, OP, and MKP within the ACO framework. Our method automatically designs the heuristic information components that influence state transitions during solution construction. Results within the ACO framework across different methods are shown in Table \ref{tab:2}. The heuristics designed by our method achieve substantial improvements over manually designed heuristics in classical ACO. It also surpasses NCO methods that use neural networks to predict heuristic information on most problems. Notably, our method exhibits competitive performance compared to existing LLM-based AHD approaches.

\paragraph{Genetic Algorithm.}

Genetic Algorithms (GA) evolve a population of solutions via selection, crossover, and mutation, making them suitable for complex optimization problems. Following experiments in ReEvo, we evaluate our method on the Decap Placement Problem (DPP), automatically designing problem-specific crossover operators to improve crossover strategies for generating new offspring in our experiment. In Table~\ref{tab:2}, we compare our results with heuristics designed by ReEvo.

\begin{table}[h]
  \centering
\caption{
Ablation study of TIDE on TSP, KP, and online BPP.
Variants remove 3 key components to evaluate their contribution, with results reported as relative gaps (\%). For KP, $W = 12.5$ for 50-item instances and $W = 25$ for 100- and 200-item instances; for online BPP, the capacity is 100. All results are averaged over three runs. The best results are highlighted in bold.
}
    \begin{tabular}{l|rrr}
    \toprule
    \multicolumn{4}{c}{TSP} \\
    \midrule
    \multicolumn{1}{c|}{Scale} & \multicolumn{1}{c}{50} & \multicolumn{1}{c}{100} & \multicolumn{1}{c}{200} \\
    \midrule
    w/o Migration Strategy & 8.00\% & 9.97\% & 10.73\% \\
    w/o UCB Selection Policy & 5.87\% & 8.91\% & 12.64\% \\
    w/o Parameter Refinement & 8.01\% & 9.74\% & 10.85\% \\
    TIDE  & \textbf{4.76\%} & \textbf{7.15\%} & \textbf{10.65\%} \\
    \midrule
    \multicolumn{4}{c}{KP} \\
    \midrule
    \multicolumn{1}{c|}{Scale} & \multicolumn{1}{c}{50} & \multicolumn{1}{c}{100} & \multicolumn{1}{c}{200} \\
    \midrule
    w/o Migration Strategy & 0.19\% & 0.09\% & 0.08\% \\
    w/o UCB Selection Policy & 0.13\% & 0.06\% & 0.05\% \\
    w/o Parameter Refinement & 0.17\% & 0.07\% & 0.07\% \\
    TIDE  & \textbf{0.04\%} & \textbf{0.02\%} & \textbf{0.02\%} \\
    \midrule
    \multicolumn{4}{c}{Online BPP } \\
    \midrule
    \multicolumn{1}{c|}{Scale} & \multicolumn{1}{c}{1k} & \multicolumn{1}{c}{5k} & \multicolumn{1}{c}{10k} \\
    \midrule
    w/o Migration Strategy & 3.58\% & 2.63\% & 2.47\% \\
    w/o UCB Selection Policy & 3.64\% & 1.70\% & 1.34\% \\
    w/o Parameter Refinement & 4.26\% & 1.83\% & 1.51\% \\
    TIDE  & \textbf{3.33\%} & \textbf{0.73\%} & \textbf{0.36\%} \\
    \bottomrule
    \end{tabular}%
  \label{tab:ablation}%
\end{table}%




\paragraph{Reinforcement Learning.}
Thanks to the strong language understanding and generality of large models, we are able to design heuristics for more complex frameworks. Following experiments in MCTS-AHD, we evaluate our method on a policy optimization task Mountain Car-v0, with a reinforcement learning framework. We compare our method with LLM-based AHD methods and demonstrate clear advantages, as shown in Table~\ref{tab:2}.

\paragraph{Ablation Study.}

To better understand the contribution of each component in the proposed method, we conduct a comprehensive ablation study by systematically removing or modifying key modules, with the results reported in Table \ref{tab:ablation}. All ablation experiments are performed under the same experimental settings as the full model to ensure fair comparison. Specifically, the first variant removes the island migration mechanism and instead performs random migration. The second variant eliminates the adaptive operator selection mechanism and resorts to random selection. The third variant removes the hyperparameter tuning module. The results demonstrate that removing any single component leads to a noticeable degradation in solution quality, indicating that each module plays a complementary role in the proposed framework.


\paragraph{Comparative Experiment.}

\begin{figure}[h]
    \centering
    \includegraphics[width=0.6\linewidth]{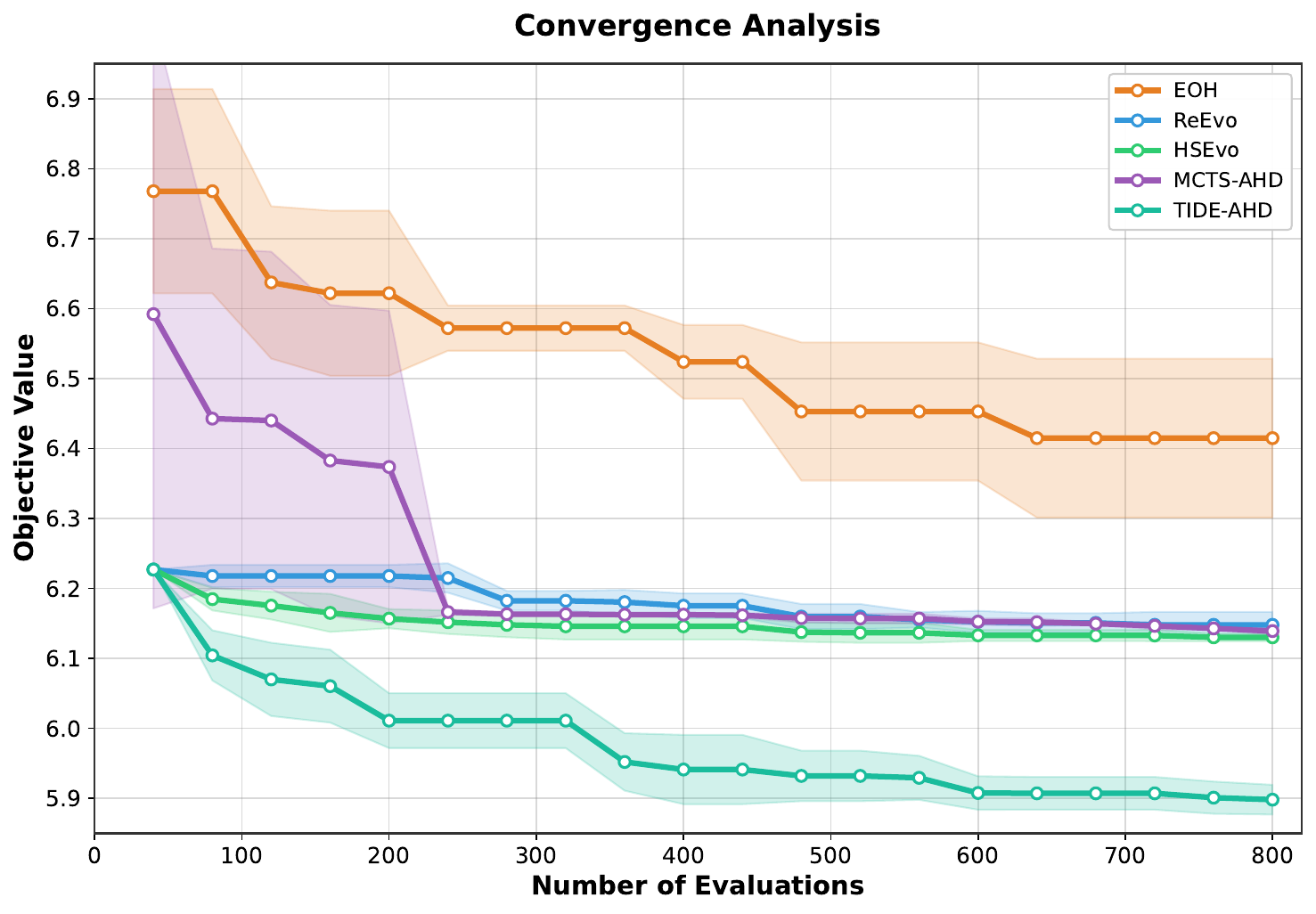}
    \caption{A key example on TSP50 under constructive framework: Convergence comparison of different LLM-based AHD methods. The x-axis denotes the number of evaluations, and the y-axis represents the objective value.}
    \label{fig:convergence}
\end{figure}

We compare the convergence behavior of different LLM-based AHD methods in terms of objective value versus the number of evaluations. As shown in Figure \ref{fig:convergence}, our method achieves better results under the same number of evaluations. Specifically, methods initialized with seed functions, such as ReEvo and HSEvo, tend to converge prematurely with limited improvement, while our method continues to improve and consistently achieves better objective values. Specifically, ReEvo, HSEvo, and our method all adopt a warm-start strategy with reflection, i.e., they are initialized with seed functions. However, since our method is island-based, it maintains richer population diversity, and the adaptive operator selection facilitates escaping local optima, thereby avoiding the premature convergence observed in other methods.

\section{Conclusion}
In this paper, we introduced TIDE to overcome the limitations of monolithic code generation in Automated Heuristic Design by employing a nested architecture that decouples algorithmic logic from continuous parameters. To achieve this, TIDE synergizes an outer TSED-guided island model for structural diversity with an inner loop that combines LLM reasoning and gradient-free tuning to mitigate numerical blindness. Extensive experiments across nine combinatorial optimization tasks demonstrate that TIDE not only discovers superior heuristics compared to state-of-the-art baselines but also improves search efficiency by minimizing token consumption on parameter tuning. Future work will explore applying TIDE to multi-objective optimization scenarios and investigating the transferability of learned structural insights across disparate problem domains.

\bibliographystyle{unsrtnat}
\bibliography{references}  
\newpage
\appendix
\onecolumn
\section{Related Work}

\subsection{LLM-Driven Automated Heuristic Design}
The paradigm of Automated Heuristic Design (AHD) has fundamentally evolved from syntax-oriented Genetic Programming (GP) to semantic-aware generation driven by Large Language Models (LLMs)~\citep{sabar2014automatic,yu2024deep}. This transition capitalizes on the ability of LLMs to act as intelligent \textit{variation operators}, generating executable code from natural language prompts rather than performing random symbolic mutations. Foundational frameworks such as FunSearch~\citep{romera2024mathematical} and EoH~\citep{liu2024evolution} established the efficacy of this approach by integrating LLMs within evolutionary loops to iteratively refine algorithms for combinatorial problems. Building on this, ReEvo~\citep{ye2024reevo} introduced reflective mechanisms, utilizing short-term and long-term memory to guide the heuristic search process explicitly.

To overcome the limitations of standard evolutionary search, recent research has diversified into more sophisticated exploration strategies. MCTS-AHD~\citep{zheng2025monte} adapts Monte Carlo Tree Search to the space of programs, enabling a more comprehensive traversal of algorithmic possibilities compared to greedy or local search methods. Parallel efforts have focused on the critical challenge of population diversity. PartEvo~\citep{hu2025partition} introduces feature-assisted partitioning to maintain distinct niches of solutions, thereby managing the exploration-exploitation trade-off. Similarly, HSEvo~\citep{dat2025hsevo} incorporates entropy-based diversity metrics within a Harmony Search framework to prevent premature convergence toward homogeneous code structures.

Beyond single-objective optimization, the field has expanded toward robustness and dynamic adaptation. Acknowledging that a single heuristic rarely generalizes across all problem distributions, EoH-S~\citep{liu2025eoh} shifts the design objective from optimizing individual algorithms to evolving complementary heuristic sets. Furthermore, moving beyond fixed LLM generators, recent methodologies have integrated learning signals directly into the generation process. Methods such as OPRO~\citep{yang2023large} leverage optimization trajectories as meta-prompts, while LLaMoCo~\citep{ma2024llamoco} and EvoTune~\citep{surina2025algorithm} employ instruction tuning and Reinforcement Learning (RL) to fine-tune the backbone model's internal weights for code generation tasks. NeRM~\citep{guo2025nested} further extends this by coupling generation with a predictor-assisted evaluation model in a nested co-evolutionary cycle.

\subsection{Island Models in Evolutionary Computation}
The Island Model represents a canonical paradigm in parallel evolutionary computation designed to mitigate premature convergence by decomposing the global population into semi-isolated subpopulations~\citep{whitley1999island,skolicki2005analysis}. This spatial structuring maintains genotypic diversity by restricting interaction to periodic migration events, thereby allowing distinct evolutionary trajectories to develop concurrently~\citep{rucinski2010impact}. Foundational research has extensively characterized the impact of communication topologies, including ring~\citep{kushida2013island,abdelhafez2019performance,wu2019ensemble,fu2020surrogate} and torus configurations~\citep{awadallah2020island,duarte2021island}, on the propagation of genetic material~\citep{wu2019ensemble}. These studies establish that the connectivity structure critically regulates the trade-off between the diffusion speed of superior traits and the preservation of local diversity. While static topologies have been widely adopted, the field has also advanced towards dynamic topology adaptation where connectivity evolves in response to search stagnation or population states~\citep{duarte2017dynamic}. In the emerging domain of Language Hyper-Heuristics, frameworks like FunSearch~\citep{romera2024mathematical} have integrated multi-island architectures to scale candidate generation~\citep{novikov2025alphaevolve,lee2025evolving}. However, current implementations in this specific domain predominantly rely on static isolation or randomized migration, often bypassing the sophisticated adaptive mechanisms developed in the broader evolutionary computation literature.

\subsection{Code Representation and Similarity}
The quantification of code similarity~\citep{ragkhitwetsagul2018comparison,novak2019source} is a foundational challenge in Software Engineering and AHD, critical for tasks ranging from clone detection~\citep{ain2019systematic} to evolutionary diversity management. Methodologies in this domain have historically evolved through three distinct paradigms. Lexical metrics, such as Levenshtein distance~\citep{zhang2017research} and Jaccard similarity~\citep{xu2018efficient,christiani2018scalable}, operate on surface-level text or token sequences~\citep{majumder2016semantic}. While computationally efficient, these metrics are notoriously sensitive to non-functional variations like formatting style and variable naming. With the advent of large-scale representation learning, semantic approaches employing pre-trained models (e.g., CodeBERT~\citep{feng2020codebert}, GraphCodeBERT~\citep{guo2020graphcodebert}, CrystalBLEU~\citep{eghbali2022crystalbleu}) have gained prominence. These methods map code snippets into dense vector spaces to capture functional intent, a technique recently adapted in heuristic generation frameworks like PartEvo~\citep{hu2025partition} for niche construction. Parallel to these, structural metrics leverage static analysis representations, particularly Abstract Syntax Trees (AST)~\citep{neamtiu2005understanding} and Control Flow Graphs (CFG)~\citep{sendjaja2021evaluating}. Measures such as Tree Edit Distance (TED)~\citep{sidorov2015computing,pawlik2015efficient,schwarz2017new} quantify topological discrepancies by calculating the minimum edit operations between tree structures~\citep{kartal2024automating}. Unlike probabilistic embeddings, structural metrics provide a deterministic basis for identifying algorithmic isomorphism and logical divergence, independent of syntactic sugar.

\subsection{Adaptive Operator Selection Strategies}
In evolutionary computation, the performance of variation operators exhibits intrinsic non-stationarity, fluctuating significantly across different stages of the search process. Adaptive Operator Selection (AOS)~\citep{fialho2010adaptive,pei2025adaptive} has thus emerged as a critical methodology to dynamically allocate computational resources by learning from historical performance. Early research focused on empirical heuristics, such as Probability Matching (PM) and Adaptive Pursuit (AP), which adjust selection probabilities proportional to recent rewards. To provide rigorous theoretical guarantees for the exploration-exploitation trade-off, the field subsequently adopted the Multi-Armed Bandit (MAB) formulation~\citep{lai1985asymptotically,garivier2008upper,li2010contextual}. Algorithms like the Upper Confidence Bound (UCB)~\citep{auer2002finite} provide deterministic bounds on regret, ensuring optimal asymptotic behavior~\citep{dacosta2008adaptive,fialho2010analyzing,modi2017qos}. These mechanisms have become foundational not only in operator scheduling for discrete optimization but also in tree-search algorithms~\citep{pei2025adaptive}, exemplified by the Upper Confidence bounds for Trees (UCT) used in Monte Carlo Tree Search (MCTS)~\citep{zheng2025monte}. In the emerging domain of LLM-based optimization, where inference incurs substantial cost, recent studies have begun investigating such adaptive mechanisms~\citep{zhang2025laos} to navigate the expansive space of prompting strategies efficiently.

\subsection{Memetic Algorithms and Hybridization}
Heuristic algorithms are inherently dualistic, comprising a discrete symbolic structure for logic flow and continuous numerical parameters for coefficients. Historically, the \textit{Memetic Algorithm} (MA) framework addressed this duality by hybridizing global evolutionary search (for structural exploration) with intensive local search (for individual refinement)~\citep{moscato1989evolution,neri2011handbook}. This separation of concerns is also evident in the field of Automated Algorithm Configuration (AAC), where specialized tools like Irace~\citep{lopez2016irace} and SMAC~\citep{hutter2011sequential} were developed specifically to tune continuous parameters for fixed algorithmic skeletons. Recent advancements in Neural Combinatorial Optimization (NCO) have similarly adopted hybrid strategies, integrating neural policy learning with gradient-free local search to enhance solution quality~\citep{ma2019combinatorial}. In the emerging domain of LLM-based algorithm design, while LLMs demonstrate strong symbolic reasoning capabilities, recent studies have begun to characterize their limitations in precise numerical regression~\citep{feng2024numerical}. Consequently, exploring hybrid architectures that synergize language-based logic generation with specialized numerical optimizers represents a growing frontier in the literature.

\section{Task and Framework Details.}

\subsection{Problem Formulations}
\label{app:problem}

We evaluate our method on a diverse set of representative NP-hard combinatorial optimization problems and control tasks. We formally define the objective functions and constraints for each problem below.
\paragraph{Traveling Salesman Problem (TSP)}
Given a complete graph $G = (V, E)$ with node set $V = \{1, \ldots, n\}$ representing $n$ cities, and a cost matrix where $c_{ij} \geq 0$ denotes the travel cost between city $i$ and city $j$, the Traveling Salesman Problem (TSP) aims to find a minimum-cost Hamiltonian cycle, i.e., a tour that visits every city exactly once before returning to the origin~\citep{lawler1985traveling}.
We define binary variables $x_{ij} \in \{0, 1\}$ to indicate whether the edge from city $i$ to city $j$ is included in the tour. The TSP can be formulated as the following integer program~\citep{hoffman2013traveling}:
\begin{align}
    \min \quad & \sum_{i \in V} \sum_{j \in V} c_{ij} x_{ij} \label{eq:tsp_obj} \\
    \text{s.t.} \quad & \sum_{j \in V} x_{ij} = 1, \quad \forall i \in V \label{eq:tsp_out} \\
    & \sum_{i \in V} x_{ij} = 1, \quad \forall j \in V \label{eq:tsp_in} \\
    & \sum_{i \in S} \sum_{j \in S} x_{ij} \leq |S| - 1, \quad \forall S \subset V, \ 2 \leq |S| \leq n-1 \label{eq:tsp_subtour} \\
    & x_{ij} \in \{0, 1\}, \quad \forall i, j \in V \label{eq:tsp_binary}
\end{align}
Constraints~\eqref{eq:tsp_out} and~\eqref{eq:tsp_in} enforce that each city has exactly one outgoing and one incoming edge, respectively. Constraint~\eqref{eq:tsp_subtour} eliminates subtours by ensuring that no proper subset of cities forms a disconnected cycle. The TSP is known to be NP-hard, making it computationally intractable to find optimal solutions for large-scale instances.

\paragraph{Capacitated Vehicle Routing Problem (CVRP)}
CVRP generalizes TSP to multiple routes serving a set of customers using a homogeneous fleet with capacity $Q$. Let $V=\{0\} \cup V_c$ be the node set, where $0$ represents the depot and $V_c=\{1, \dots, N\}$ represents $N$ customer nodes. Each customer $i \in V_c$ has a deterministic demand $\delta_i$.
A solution is a set of routes $\mathcal{R} = \{r_1, \dots, r_K\}$, where each route $r_k = (v_0^k, v_1^k, \dots, v_{m_k}^k, v_{m_k+1}^k)$ starts and ends at the depot (i.e., $v_0^k = v_{m_k+1}^k = 0$)~\citep{toth2002models}. The objective is to minimize the total travel distance subject to capacity constraints:
\begin{equation}
    \min_{\mathcal{R}} \sum_{k=1}^{K} \sum_{t=0}^{m_k} d_{v_t^k, v_{t+1}^k} \quad \text{s.t.} \quad \sum_{t=1}^{m_k} \delta_{v_t^k} \leq Q, \quad \forall k \in \{1, \dots, K\},
\end{equation}
ensuring that every customer $i \in V_c$ is visited exactly once across all routes.

\paragraph{Orienteering Problem (OP)}
The Orienteering Problem~\citep{vansteenwegen2011orienteering} combines elements of Knapsack and TSP. Given a set of nodes $V = \{0, 1, \dots, N\}$ in a metric space, where node $0$ is the depot. Each node $i$ is associated with a prize $p_i \geq 0$ (with $p_0=0$). Let $d_{ij}$ denote the travel cost between node $i$ and $j$. The objective is to find a tour $\tau = (v_1, \dots, v_k)$ starting and ending at the depot ($v_1=v_k=0$) that maximizes the total collected prize, subject to a maximum travel budget $T_{max}$:
\begin{equation}
    \max_{\tau} \sum_{i \in \text{set}(\tau)} p_i \quad \text{s.t.} \quad \sum_{t=1}^{k-1} d_{v_t, v_{t+1}} \leq T_{max},
\end{equation}
where $\text{set}(\tau)$ denotes the set of unique nodes visited in the tour. Each node can be visited at most once.

\paragraph{0-1 Knapsack Problem (KP)}
Consider a set of $N$ items, where item $i$ is associated with a value $v_i$ and a weight $w_i$. Given a knapsack with capacity $W$, the problem seeks a binary selection vector $\mathbf{x} \in \{0, 1\}^N$ to maximize the total value without exceeding the weight limit~\citep{zhou2023nature}:
\begin{equation}
    \max_{\mathbf{x}} \sum_{i=1}^{N} v_i x_i \quad \text{s.t.} \quad \sum_{i=1}^{N} w_i x_i \leq W, \quad x_i \in \{0, 1\}.
\end{equation}

\paragraph{Multidimensional Knapsack Problem (MKP)}
MKP extends KP by introducing $M$ distinct resource constraints~\citep{puchinger2010multidimensional}. Each item $j \in \{1, \dots, N\}$ has a value $v_j$ and consumes $w_{ij}$ units of the $i$-th resource ($i \in \{1, \dots, M\}$). Let $C_i$ denote the capacity of the $i$-th dimension. The objective is to maximize total value subject to all resource constraints:
\begin{equation}
    \max_{\mathbf{x}} \sum_{j=1}^{N} v_j x_j \quad \text{s.t.} \quad \sum_{j=1}^{N} w_{ij} x_j \leq C_i, \quad \forall i \in \{1, \dots, M\},
\end{equation}
where $x_j \in \{0, 1\}$ is the binary decision variable for item $j$.

\paragraph{Bin Packing Problem (BPP)}
Given a set of items $\mathcal{I} = \{1, \dots, N\}$ with sizes $s_i \in (0, C]$, the goal is to partition $\mathcal{I}$ into a minimum number of disjoint subsets (bins) $B_1, \dots, B_K$ such that the sum of sizes in each bin does not exceed the capacity $C$~\citep{munien2021metaheuristic}.
\begin{itemize}
    \item \textbf{Offline BPP:} The solver has full access to the set of items $\mathcal{I}$ and their sizes \textit{a priori}~\citep{borgulya2021hybrid}.
    \item \textbf{Online BPP:} Items arrive sequentially. At step $t$, item $i_t$ must be irreversibly assigned to a bin before observing $i_{t+1}$~\citep{seiden2002online}.
\end{itemize}
Mathematically, we minimize $K$ subject to $\sum_{i \in B_k} s_i \leq C$ for all $k \in \{1, \dots, K\}$ and $\bigcup_{k} B_k = \mathcal{I}$.

\paragraph{Admissible Set Problem (ASP)}
ASP is a combinatorial design problem that constructs a set of vectors $\mathcal{A} \subset \{0, 1, 2\}^n$ with maximal cardinality~\citep{barwise2017admissible}. Each vector $\mathbf{a} \in \mathcal{A}$ must have a fixed Hamming weight $\|\mathbf{a}\|_0 = w$. Furthermore, for any distinct triplet $\mathbf{a}, \mathbf{b}, \mathbf{c} \in \mathcal{A}$, there must exist a coordinate index $k$ such that the values $\{a_k, b_k, c_k\}$ form a valid configuration set $S_{valid} = \{\{0, 0, 1\}, \{0, 0, 2\}, \{0, 1, 2\}\}$ (modulo permutation). The formulation is:
\begin{equation}
    \max_{\mathcal{A}} |\mathcal{A}| \quad \text{s.t.} \quad \forall \mathbf{a} \in \mathcal{A}, \|\mathbf{a}\|_0 = w; \quad \forall \text{distinct } \mathbf{a, b, c} \in \mathcal{A}, \exists k : \{a_k, b_k, c_k\} \in S_{valid}.
\end{equation}

\paragraph{Decap Placement Problem (DPP)}
DPP addresses the optimization of Power Distribution Networks (PDNs)~\citep{smith1999power}. It is formulated as a black-box combinatorial optimization problem on a 2D discrete grid $\mathcal{G} = \{(i, j) \mid 1 \leq i, j \leq L\}$.
A problem instance is defined by a context $\mathbf{c} = (p, \mathcal{K})$, where $p \in \mathcal{G}$ is the probing port location and $\mathcal{K} \subset \mathcal{G}$ denotes keep-out regions. The objective is to select a binary placement matrix $\mathbf{x} \in \{0, 1\}^{L \times L}$ to minimize the PDN impedance, subject to a maximum capacitor count $N_{max}$:
\begin{equation}
    \min_{\mathbf{x}} \Phi(\mathbf{x}; \mathbf{c}) \quad \text{s.t.} \quad \sum_{(i,j) \in \mathcal{G}} x_{ij} \leq N_{max}, \quad \text{and} \quad x_{ij} = 0, \forall (i,j) \in \mathcal{K} \cup \{p\},
\end{equation}
where $\Phi(\cdot)$ is a black-box oracle (e.g., SPICE simulation) evaluating the maximum impedance over a target frequency range.

\paragraph{Mountain Car Control (MountainCar-v0)}
MountainCar-v0 is a classic continuous control benchmark involving an underpowered car driving up a steep hill. The state $s_t = (x_t, v_t) \in \mathcal{S}$ represents position and velocity. The discrete action space is $\mathcal{A} = \{0, 1, 2\}$. The dynamics follow a deterministic transition rule $s_{t+1} = f(s_t, a_t)$.
The objective is to synthesize a heuristic policy $\pi(\cdot)$ that minimizes the trajectory length $L$ required to reach the goal state $x_{goal} \ge 0.5$:
\begin{equation}
    \min_{\pi} L \quad \text{s.t.} \quad x_L \ge 0.5, \quad s_{t+1} = f(s_t, \pi(s_t)), \quad s_0 \in \mathcal{S}_{init},
\end{equation}
where $\mathcal{S}_{init}$ is the distribution of initial states (typically $x_0 \in [-0.6, -0.4], v_0 = 0$), and $L$ is bounded by a maximum horizon $T_{max}=200$.~\citep{zheng2025monte}

\subsection{General Optimization Frameworks}
\label{app:frameworks}

We evaluate our method by designing key heuristic functions within three established algorithmic frameworks: \textit{Constructive Heuristics}, \textit{Ant Colony Optimization} (ACO), and \textit{Local Search} variants. This demonstrates the framework-agnostic capability of our approach. In each setting, the framework provides the high-level search logic (the skeleton), while our method optimizes a specific parameterized function or heuristic component (the brain).

\subsubsection{Constructive Heuristic}
\label{app:constructive}

Constructive heuristics generate a solution by sequentially selecting discrete decision variables until a valid complete solution is formed~\citep{marti2010heuristic,cormen2022introduction}. Let $S_t$ be the partial solution at step $t$, and $\mathcal{A}_t$ be the set of available candidate actions (e.g., unvisited cities or unpacked items). A policy function $h: \mathcal{A}_t \times S_t \to \mathbb{R}$ assigns a priority score to each candidate. The algorithm deterministically selects the action with the highest score:
\begin{equation}
    a^* = \operatorname{argmax}_{a \in \mathcal{A}_t} h(a, S_t).
\end{equation}
We employ this framework for the following problems:
\begin{itemize}
    \item \textbf{TSP:} The heuristic selects the next node to visit. The input state includes the current node location, the set of unvisited nodes, and the distance matrix~\citep{rosenkrantz1977analysis,applegate2011traveling}.
    \item \textbf{KP:} The heuristic ranks items to be added to the knapsack. Inputs include the value and weight of remaining items and the residual capacity~\citep{assi2018survey}.
    \item \textbf{ASP:} The heuristic evaluates an $n$-dimensional vector's suitability for inclusion in the admissible set $\mathcal{A}$, ensuring constraint satisfaction~\citep{ellenberg2017large,romera2024mathematical}.
    \item \textbf{Online BPP:} Upon the arrival of a new item, the heuristic assigns a preference score to each existing bin (or a new bin) based on the item's size and the bins' residual capacities~\citep{coffman2013bin}.
\end{itemize}

\subsubsection{Improvement Heuristic}
\label{app:local_search}

Local search algorithms iteratively improve a candidate solution by exploring its neighborhood. To overcome the limitation of getting trapped in local optima, we employ meta-heuristic frameworks that incorporate guidance mechanisms~\citep{peres2021combinatorial,tsai2023handbook}. We apply our method to automate the core heuristic functions within two distinct frameworks: \textit{Guided Local Search} (GLS) and \textit{Knowledge-Guided Local Search} (KGLS).

\paragraph{Guided Local Search}
Standard GLS~\citep{voudouris2025guided} escapes local optima by dynamically augmenting the objective function with penalty terms associated with solution features (e.g., edges in a tour). Let $f(s)$ be the original cost of a solution $s$. GLS minimizes an augmented cost function $g(s)$:
\begin{equation}
    g(s) = f(s) + \lambda \cdot \sum_{(i,j) \in E} p_{ij} \cdot \mathbb{I}_{ij}(s),
\end{equation}
where $E$ is the set of all edges, $p_{ij}$ represents the accumulated penalty count for edge $(i, j)$, and $\mathbb{I}_{ij}(s)$ is an indicator function that equals 1 if edge $(i, j)$ is present in solution $s$, and 0 otherwise. $\lambda$ is a regularization parameter.

The core heuristic logic lies in determining \textit{which} feature to penalize when the search is trapped in a local optimum $s^*$. This is governed by a utility function $\mu(i, j)$. Standard GLS defines this utility as $\mu(i, j) = d_{ij} / (1 + p_{ij})$. In our framework, we task the LLM to evolve a more effective utility function $\Psi_{\text{GLS}}$:
\begin{equation}
    \mu(i, j) = \Psi_{\text{GLS}}(d_{ij}, p_{ij}, \dots).
\end{equation}
The edge with the maximum utility in $s^*$ is penalized ($p_{ij} \leftarrow p_{ij} + 1$), altering the landscape to guide the solver away from the current basin of attraction.

\paragraph{Knowledge-Guided Local Search}
We also evaluate the Knowledge-Guided Local Search framework~\citep{arnold2019knowledge}, which operates as a specialized \textit{Iterated Local Search} (ILS)~\citep{lourencco2018iterated}. Unlike GLS, KGLS does not modify the objective function during the local descent. Instead, it utilizes a pre-computed guidance matrix to bias the perturbation phase.

The algorithm alternates between two stages:
\begin{enumerate}
    \item \textbf{Local Descent:} A standard descent algorithm (e.g., 2-opt~\citep{karagul2016using}, Relocate) minimizes the original cost $f(s)$ based on the distance matrix $\mathbf{D}$ until a local optimum is reached.
    \item \textbf{Guided Perturbation:} To escape the local optimum, the solution is perturbed. This perturbation is biased by a guidance matrix (or heatmap) $\bm{\Omega} \in \mathbb{R}^{N \times N}$.
\end{enumerate}

Our method automates the design of the mapping function $\Psi_{\text{KGLS}}$ that constructs this matrix from the raw problem instance $\mathcal{I}$ (e.g., coordinates and distances):
\begin{equation}
    \bm{\Omega} = \Psi_{\text{KGLS}}(\mathcal{I}).
\end{equation}
The matrix $\bm{\Omega}$ encodes the global desirability of edges. During perturbation, edges with higher values in $\bm{\Omega}$ are preferentially inserted, while edges with lower values are removed, guiding the solver toward globally promising regions of the search space.

\subsubsection{Ant Colony Optimization}
\label{app:aco}

ACO is a meta-heuristic that constructs solutions probabilistically based on dynamic pheromone trails ($\tau$) and static heuristic information ($\eta$)~\citep{dorigo2018ant,dorigo2018introduction}. The probability of selecting a component $e_{ij}$ (e.g., an edge moving from node $i$ to $j$) is typically proportional to $\tau_{ij}^\alpha \cdot \eta_{ij}^\beta$. While $\tau$ evolves dynamically during the search based on solution quality, the heuristic information $\eta$ is traditionally a fixed matrix derived from problem features (e.g., $\eta_{ij} = 1/d_{ij}$ in standard TSP).

Our method aims to automate the design of the function that generates this heuristic matrix $\bm{\eta}$. We adopt the DeepACO framework~\citep{ye2023deepaco}, where the learned function maps problem instances to an $N \times N$ matrix $\bm{\eta}$. We apply this framework to the following problem settings:

\begin{itemize}
    \item \textbf{Routing Problems (TSP, CVRP, OP):} For TSP and CVRP, the heuristic function maps spatial features (distances) and node constraints (demands) to an edge desirability matrix~\citep{applegate2011traveling,toth2014vehicle}. For the Orienteering Problem (OP), the heuristic explicitly incorporates node prizes to balance the trade-off between collecting rewards and conserving the travel budget~\citep{gunawan2016orienteering}.
    \item \textbf{Packing Problems (MKP, Offline BPP):} The function maps item-bin compatibility features to a preference matrix, prioritizing efficient packing configurations to minimize resource usage~\citep{coffman2013bin}.
\end{itemize}

\subsubsection{Genetic Algorithm}
\label{app:ga}

Genetic Algorithms (GA) are population-based meta-heuristics that evolve a set of candidate solutions over multiple generations, inspired by natural selection~\citep{holland1992adaptation}. Unlike the constructive or local search policies described above, which operate on a single solution trajectory, GAs maintain a population $\mathcal{P}_t = \{\mathbf{x}_1, \dots, \mathbf{x}_P\}$ at generation $t$. 

In the context of the Decap Placement Problem (DPP)~\citep{ye2024reevo}, each $\mathbf{x}_i$ represents a discrete set of placement locations. The evolutionary cycle typically involves three stages~\citep{eiben2015introduction}:
\begin{enumerate}
    \item \textbf{Selection:} A subset of high-performing individuals (elites) is preserved based on the objective function value (e.g., power integrity reward). Parents are selected from $\mathcal{P}_t$ to produce offspring.
    \item \textbf{Variation (Crossover \& Mutation):} New solutions are generated by recombining information from parents and introducing stochastic perturbations.
    \item \textbf{Repair/Validation:} Domain-specific constraints (e.g., keep-out regions, non-overlapping indices) are enforced to ensure solution feasibility.
\end{enumerate}

While the high-level GA loop (initialization, selection, replacement) is fixed, our method automates the design of the core variation operators, aligning with the principles of generative hyper-heuristics~\citep{pillay2018hyper}. We define two heuristic functions to be evolved:
\begin{itemize}
    \item \textbf{Crossover Operator ($\Psi_{\text{cross}}$):} A function that takes a set of parent solutions and produces offspring by combining their substructures.
    $$ \mathbf{x}_{\text{child}} = \Psi_{\text{cross}}(\mathbf{x}_{\text{parent1}}, \mathbf{x}_{\text{parent2}}) $$
    \item \textbf{Mutation Operator ($\Psi_{\text{mut}}$):} A function that introduces diversity by modifying an individual with a specific probability.
    $$ \mathbf{x}_{\text{new}} = \Psi_{\text{mut}}(\mathbf{x}_{\text{child}}) $$
\end{itemize}
Our method searches for the optimal logic within $\Psi_{\text{cross}}$ and $\Psi_{\text{mut}}$ that balances exploration (diversity) and exploitation (preserving good placement patterns) specifically for the EDA landscape.

\subsubsection{Reinforcement Learning}
\label{app:rl}

Reinforcement Learning (RL) addresses sequential decision-making problems where an agent interacts with a dynamic environment to maximize cumulative rewards~\citep{barto2021reinforcement}. Unlike the static combinatorial problems discussed previously, RL tasks involve continuous state spaces and long-term temporal dependencies. In this framework, the goal of Automated Heuristic Design is to synthesize a \textit{Policy Function} $\pi: \mathcal{S} \to \mathcal{A}$, which maps the current state $s_t$ directly to an optimal action $a_t$~\citep{liang2022code}.

We evaluate this framework on the classic control task \textit{MountainCar-v0}, originally formulated by \citep{moore1990efficient} and standardized in the OpenAI Gym suite~\citep{brockman2016openai}. The problem is characterized by an underpowered vehicle positioned between two mountains. The engine lacks sufficient power to climb the slope directly in a single pass. Consequently, the optimal policy requires a momentum-based strategy, where the agent must oscillate back and forth to build up sufficient potential energy to reach the target on the right hilltop.

In our experiments, the LLM is tasked with designing an explicit, interpretable heuristic function \textit{choose\_action} to serve as the agent's policy.
\begin{itemize}
    \item \textbf{State Space:} The input is a continuous vector $s = (x, \dot{x})$, representing the car's horizontal position and current velocity.
    \item \textbf{Action Space:} The output is a discrete action $a \in \{0, 1, 2\}$, corresponding to pushing left, applying no force, or pushing right.
    \item \textbf{Design Objective:} The generated heuristic must implement logical rules (e.g., threshold-based decisions or phase-space analysis) to determine the action that minimizes the number of steps required to reach the goal state ($x \ge 0.5$).
\end{itemize}
Unlike Deep Reinforcement Learning methods that encode policies in opaque neural networks~\citep{mnih2015human, schulman2017proximal}, this approach aims to discover symbolic control laws that are both effective and human-readable~\citep{trivedi2021learning}.


\section{Experimental Configuration}

\subsection{Hyperparameter Settings}
\label{app:setting}
To ensure reproducibility, we detail the other specific hyperparameter configurations used in our main experiments. Table~\ref{tab:hyperparams} summarizes the default settings.
\begin{table}[h]
    \centering
    \caption{Default Hyperparameter Configuration for TIDE.}
    \label{tab:hyperparams}
    \begin{tabular}{l c c}
    \toprule
    \textbf{Hyperparameter} & \textbf{Symbol} & \textbf{Value} \\
    \midrule
    \textit{Outer Loop (Island Model)} & & \\
    \quad Island Count & $N_{\text{island}}$ & 6 \\
    \quad Population Size per Island & $N_{\text{pop}}$ & 8 \\
    \quad TSED Diversity Threshold & $\tau$ & 0.7 \\
    \quad Stagnation Tolerance Iteration (Reset) & $I_{\text{stag}}$ & 8 \\
    \quad Migration Cooldown Interval & $I_{\text{cool}}$ & 2 \\
    \midrule
    \textit{Inner Loop (Co-Evolution)} & & \\
    \quad UCB Exploration Constant & $C$ & $\sqrt{2}$ \\
    \quad Parameter Tuning Candidates & $N_{\text{tune}}$ & 3 \\
    \midrule
    \textit{LLM Generation} & & \\
    \quad Sampling Temperature & $T_{\text{LLM}}$ & 1.0 \\
    \bottomrule
    \end{tabular}
\end{table}

where $N_{\text{island}} = 6$ and $N_{\text{pop}} = 8$ define the global topology, selected to maximize parallel evolutionary trajectories while constraining the per-generation token cost.
Regarding migration dynamics, consistent with the asynchronous triggering mechanism described in Section ~\ref{sec:migration}, TIDE does not employ a fixed migration frequency. Instead, the operational cycle is regulated by $I_{\text{cool}} = 2$, which establishes a minimum temporal bound between consecutive events. This mandatory cooldown prevents oscillatory behavior and ensures local populations have sufficient generations to stabilize and assimilate new elite code or insights before further migration can occur.

$\tau = 0.7$ serves as the critical TSED threshold governing Dual-Mode Migration: pairs with similarity $S_{\text{TSED}} > \tau$ execute \textit{Code Transfer} for local exploitation, while those with $S_{\text{TSED}} \le \tau$ trigger \textit{Insight Transfer} for global exploration. $I_{\text{stag}} = 8$ sets the tolerance for the Selective Reset mechanism to resolve deep stagnation. $C = \sqrt{2}$ is adopted from standard UCB1~\citep{auer2002finite} literature to balance operator exploration with exploitation. $N_{\text{tune}} = 3$ is set as the minimal viable batch size for the parameter refinement module; this value corresponds to the theoretical lower bound required to construct a valid difference vector for mutation (i.e., one target vector plus two difference vectors), ensuring gradient-free tuning with minimal computational overhead. Finally, $T_{\text{LLM}} = 1.0$ is maintained to encourage stochastic diversity in structural reasoning, deviating from the lower temperatures typically used for precision-oriented code generation.


\begin{table}[b]
  \centering
  \caption{Parameter Sensitivity Analysis on Online BPP and Constructive KP. This table reports the performance under varying hyperparameter settings. For the island configuration, values are reported in the form of \emph{number of islands}$\times$\emph{population size per island}. The default configuration is marked in the table. Best results under each setting are highlighted in bold.}
    \begin{tabular}{cl|ccc|ccc}
    \toprule
    \multicolumn{2}{c|}{\multirow{2}[2]{*}{Parameters}} & \multicolumn{3}{c|}{BPP online} & \multicolumn{3}{c}{KP constructive} \\
    \multicolumn{2}{c|}{} & 1k    & 5k    & 10k   & 50    & 100   & 200 \\
    \midrule
    \multirow{3}[2]{*}{$C$ in UCB} & 0.5   & 419.8 & 2063.5 & 4084.1 & 20.00 & 40.23 & 57.37 \\
          & $\sqrt{2}$ (default) & 415.8 & \textbf{2034.2} & \textbf{4025.1} & \textbf{20.03} & \textbf{40.26} & \textbf{57.44} \\
          & 2     & \textbf{415.2} & 2047.3 & 4052.1 & 20.01 & 40.25 & 57.42 \\
    \midrule
    \multirow{3}[2]{*}{$N_{\text{island}} \times N_{\text{pop}}$} & 3$\times$16  & 417.2 & 2036.9 & 4032.8 & 20.00 & 40.24 & 57.41 \\
          & 6$\times$8 (default) & 415.8 & \textbf{2034.2} & \textbf{4025.1} & \textbf{20.03} & \textbf{40.26} & \textbf{57.44} \\
          & 12$\times$4  & \textbf{415.2} & 2037.7 & 4034.8 & 20.02 & 40.25 & 57.42 \\
    \midrule
    \multirow{3}[2]{*}{TSED diversity threshold ($\tau$)} & 0.5   & \textbf{413.5} & 2045.4 & 4050.7 & 20.02 & 40.25 & 57.42 \\
          & 0.7 (default) & 415.8 & \textbf{2034.2} & \textbf{4025.1} & \textbf{20.03} & \textbf{40.26} & \textbf{57.44} \\
          & 0.9   & 418.7 & 2084.4 & 4129.9 & 20.00 & 40.23 & 57.40 \\
    \bottomrule
    \end{tabular}%
  \label{tab:parameter}%
\end{table}%

\subsection{Parameter Sensitivity Analysis}

We conduct a parameter sensitivity analysis to study the impact of key hyperparameters in our framework, including the exploration coefficient $c$ in UCB, the island configuration, and the diversity threshold. The analysis is performed on two representative tasks: online BPP and constructive KP, covering different problem scales.

As shown in Table \ref{tab:parameter}, we systematically evaluate three critical hyperparameters to validate the framework's robustness. First, the UCB exploration constant $c$ determines the weight of the uncertainty term in the bandit selection policy, directly controlling the trade-off between exploring under-sampled prompt strategies and exploiting high-yield ones. The results indicate that deviating from the theoretical standard of $\sqrt{2}$ degrades search efficiency: a lower value ($c=0.5$) causes the scheduler to greedily lock onto local optima, while a higher value ($c=2.0$) results in inefficient random search behavior. 

Second, the island topology configuration defines the granularity of the global population structure. We observe that extreme fragmentation ($12 \times 4$) is detrimental because the insufficient genetic material within small local populations leads to rapid genetic drift, rendering LLM-based crossover operators ineffective. Conversely, a coarse-grained topology ($3 \times 16$) limits the global search breadth, covering fewer distinct basins of attraction. 

Third, the diversity threshold $\tau$ acts as the decision boundary for the dual-mode migration strategy. Setting $\tau$ too low ($0.5$) triggers excessive code overwriting that destroys global structural diversity, whereas setting it too high ($0.9$) hinders the necessary exploitation of elite solutions. 

Ultimately, these experiments demonstrate that TIDE maintains robust performance across reasonable parameter ranges, confirming that the default configuration effectively captures the intended architectural balance without requiring intensive user-side tuning.

\subsection{Details of Evaluation Datasets}

For all LLM-based AHD methods considered in our experiments, we use the same training and test datasets to ensure a fair and controlled comparison. All methods are trained and evaluated on identical instance sets, eliminating potential performance differences caused by data variations rather than heuristic design. The detailed numbers of instances in the training and test sets for each problem are summarized in Table \ref{tab:dataset}.

\begin{table}[t]
  \centering
  \caption{Summary of training and test datasets used across different problems and algorithmic frameworks. For each entry, the number of instances is reported, followed by key problem parameters in parentheses (e.g., number of nodes/items, capacity constraints).}
    \begin{tabular}{c|p{6.8em}p{10em}p{9.5em}c}
    \toprule
    Framework & \multicolumn{4}{c}{Constructive Heuristic} \\
    \midrule
    Problems & \multicolumn{1}{c}{TSP} & \multicolumn{1}{c}{KP} & \multicolumn{1}{c}{BPP online (WeiBull)} & ASP \\
    \midrule
    Training Datasets & 64 (50 nodes) & 64 (100 items, W=25) & 1 (1k items, W=100)\newline{}1 (1k items, W=500)\newline{}1 (5k items, W=100)\newline{}1 (5k items, W=500) & \multicolumn{1}{p{9.75em}}{1 (n=15, w=10)} \\
    \midrule
    Test Datasets & 1000 (50 nodes)\newline{}128 (100 nodes)\newline{}64 (200 nodes) & 1000 (50 items, W=12.5)\newline{}1000 (100 items, W=25)\newline{}1000 (200 items, W=25) & 5 (1k items, W=100)\newline{}5 (1k items, W=500)\newline{}5 (5k items, W=100)\newline{}5 (5k items, W=500)\newline{}5 (10k items, W=100)\newline{}5 (10k items, W=500) & \multicolumn{1}{p{9.75em}}{1 (n=12, w=7)\newline{}1 (n=15, w=10)\newline{}1 (n=21, w=15)} \\
    \midrule
    Framework & \multicolumn{2}{c}{Improvement Heuristic} & \multicolumn{2}{c}{ACO} \\
    \midrule
    Problems & \multicolumn{1}{c}{TSP GLS} & \multicolumn{1}{c}{TSP KGLS} & \multicolumn{1}{c}{TSP} & CVRP \\
    \midrule
    Training Datasets & 64 (100 nodes) & 10 (200 nodes) & 5 (50 nodes) & \multicolumn{1}{p{9.75em}}{10 (50 nodes, C=50)} \\
    \midrule
    Test Datasets & 1000 (100 nodes)\newline{}128 (200 nodes)\newline{}128 (500 nodes) & 1000 (100 nodes)\newline{}1000 (200 nodes)\newline{}64 (500 nodes) & 64 (50 nodes)\newline{}64 (100 nodes) & \multicolumn{1}{p{9.75em}}{64 (50 nodes, C=50)\newline{}64 (100 nodes, C=50)} \\
    \midrule
    Framework & \multicolumn{3}{c}{ACO} & GA \\
    \midrule
    Problems & \multicolumn{1}{c}{OP} & \multicolumn{1}{c}{MKP} & \multicolumn{1}{c}{BPP offline} & DPP \\
    \midrule
    Training Datasets & 5 (50 nodes) & 5 (100 items, m=5) & 5 (500 items, W=150) & 2 \\
    \midrule
    Test Datasets & 64 (50 nodes)\newline{}64 (100 nodes) & 64 (100 items, m=5)\newline{}64 (200 items, m=5) & 64 (500 items, W=150)\newline{}64 (1000 items, W=150) & 64 \\
    \bottomrule
    \end{tabular}%
  \label{tab:dataset}%
\end{table}%



\section{More Experimental Results}

\subsection{Results on TSPLib}
\label{app:tsplib}
TSPLIB is a widely used benchmark suite for the Traveling Salesman Problem, comprising instances with diverse sizes and heterogeneous spatial distributions, including random, clustered, and structured layouts derived from real-world scenarios.

To further demonstrate the robustness and generalization capability of the heuristics designed by our method, we conduct additional experiments on TSPLIB instances. Compared to synthetically generated data, TSPLIB provides a more challenging and diverse testbed, enabling a comprehensive evaluation of heuristic performance across varying instance scales and distribution patterns. These experiments allow us to assess whether the proposed heuristics can consistently deliver improvements under different structural characteristics of problem instances.

\begin{table}[htbp]
  \centering
  \caption{Results on various TSPLib instances. The best results are highlighted in bold.}
    \begin{tabular}{lccccc}
    \toprule
    \multicolumn{1}{c}{name} & EoH   & ReEvo & HSEvo & MCTS-AHD & TIDE \\
    \midrule
    eil51 & 14.07\% & 8.76\% & 9.25\% & 13.36\% & \textbf{5.16\%} \\
    st70  & 13.19\% & 12.77\% & 10.48\% & 9.81\% & \textbf{6.44\%} \\
    eil76 & 15.58\% & 10.52\% & 9.82\% & 11.33\% & \textbf{9.78\%} \\
    pr76  & 17.71\% & 14.25\% & 11.46\% & 16.61\% & \textbf{9.76\%} \\
    rat99 & 18.30\% & 13.37\% & 12.77\% & 10.49\% & \textbf{8.01\%} \\
    kroA100 & 15.20\% & 11.79\% & 10.12\% & 11.38\% & \textbf{7.05\%} \\
    kroB100 & 16.33\% & 10.70\% & 12.05\% & 10.50\% & \textbf{5.89\%} \\
    kroC100 & 14.47\% & 12.64\% & 12.36\% & 8.49\% & \textbf{7.97\%} \\
    kroD100 & 19.67\% & 11.08\% & 15.38\% & 12.15\% & \textbf{9.39\%} \\
    kroE100 & 18.88\% & 12.34\% & 13.00\% & 8.59\% & \textbf{6.59\%} \\
    rd100 & 17.40\% & 12.63\% & 14.00\% & 9.88\% & \textbf{9.27\%} \\
    eil101 & 20.96\% & 11.77\% & 13.12\% & 14.60\% & \textbf{9.69\%} \\
    pr107 & 4.41\% & 6.14\% & 6.58\% & 2.49\% & \textbf{1.61\%} \\
    pr124 & 14.53\% & 15.93\% & 13.27\% & 5.50\% & \textbf{5.11\%} \\
    pr144 & 6.58\% & 7.46\% & 9.18\% & 3.76\% & \textbf{3.65\%} \\
    ch150 & 11.28\% & 10.35\% & 10.35\% & 7.66\% & \textbf{6.71\%} \\
    kroA150 & 17.69\% & 12.00\% & 13.36\% & 10.71\% & \textbf{10.30\%} \\
    kroB150 & 16.91\% & 11.95\% & 11.05\% & 10.84\% & \textbf{8.46\%} \\
    pr152 & 12.01\% & 12.50\% & 11.11\% & \textbf{6.19\%} & 8.20\% \\
    u159  & 18.41\% & \textbf{10.46\%} & 13.03\% & 15.03\% & 11.96\% \\
    rat195 & 15.56\% & 8.07\% & 9.64\% & \textbf{8.02\%} & 8.43\% \\
    kroA200 & 20.41\% & 12.88\% & 12.80\% & 11.30\% & \textbf{8.04\%} \\
    kroB200 & 18.14\% & 14.61\% & 15.16\% & 13.03\% & \textbf{12.93\%} \\
    ts225 & 15.03\% & 7.89\% & \textbf{7.08\%} & 11.83\% & 11.67\% \\
    tsp225 & 18.33\% & 11.55\% & 12.62\% & 13.12\% & \textbf{9.05\%} \\
    pr226 & 14.73\% & 14.38\% & 15.49\% & 9.78\% & \textbf{8.90\%} \\
    lin318 & 19.35\% & 14.13\% & 15.04\% & 14.44\% & \textbf{13.22\%} \\
    rd400 & 17.65\% & 12.88\% & 14.36\% & \textbf{12.42\%} & 14.07\% \\
    fl417 & 17.96\% & 19.58\% & 18.15\% & 13.35\% & \textbf{11.98\%} \\
    p654  & 24.19\% & 17.56\% & 19.26\% & 17.10\% & \textbf{15.71\%} \\
    \midrule
    avg.  & 16.16\% & 12.10\% & 12.38\% & 10.79\% & \textbf{8.83\%} \\
    \bottomrule
    \end{tabular}%
  \label{tab:tsplib}%
\end{table}%

We conduct experiments on TSPLIB under the constructive framework, with results reported in Table \ref{tab:tsplib}. Our method outperforms existing LLM-based AHD approaches on most instances, indicating its effectiveness in constructing high-quality solutions across diverse problem settings.

\subsection{Results on Weibull BPP Instances}
\label{app:bpp}
In addition to the results on online BPP presented in the main paper, we further evaluate our method on broader Weibull-distributed BPP instances. Compared with uniformly distributed item sizes, Weibull distributions more closely resemble real-world packing scenarios, and these instances are generated following the standard protocols adopted in prior LLM-based AHD studies.

We construct a richer benchmark by considering a wider range of bin capacities and problem scales to comprehensively assess the robustness of the proposed heuristics. This design introduces greater diversity and difficulty into the evaluation setting, enabling a more thorough examination of performance across heterogeneous conditions.

As reported in Table \ref{tab:bpp}, our method consistently achieves lower average optimality gaps with respect to the optimal solutions across all evaluated datasets when compared to other LLM-based AHD methods. Notably, the heuristics produced by our approach also outperform traditional expert-designed heuristics, demonstrating the effectiveness of LLM-guided heuristic discovery in capturing complex packing patterns beyond manually crafted rules.

\begin{table}[htbp]
  \centering
  \caption{Results on Online BPP in Weibull instances with various capacities and problem sizes. Results marked with * denote values taken from MCTS-AHD~\cite{zheng2025monte}.}
    \begin{tabular}{cc|cc|ccccc}
    \toprule
    Capacity & Size  & First Fit* & Best Fit* & EoH & ReEvo & HSEvo & MCTS-AHD & TIDE(ours) \\
    \midrule
    \multirow{3}[2]{*}{100} & 1k    & 4.77\% & 5.02\% & 3.93\% & 4.01\% & 4.74\% & 3.76\% & 3.33\% \\
          & 5k    & 4.31\% & 4.65\% & 1.20\% & 3.40\% & 4.22\% & 2.49\% & 0.73\% \\
          & 10k   & 4.05\% & 4.36\% & 0.63\% & 3.22\% & 4.03\% & 2.13\% & 0.36\% \\
    \midrule
    \multirow{3}[2]{*}{500} & 1k    & 0.25\% & 0.25\% & 0.41\% & 0.17\% & 0.25\% & 0.41\% & 0.33\% \\
          & 5k    & 0.55\% & 0.55\% & 0.48\% & 0.46\% & 0.55\% & 0.41\% & 0.51\% \\
          & 10k   & 0.47\% & 0.50\% & 0.46\% & 0.41\% & 0.47\% & 0.37\% & 0.49\% \\
    \midrule
    \multicolumn{2}{c|}{avg.} & 2.40\% & 2.56\% & 1.18\% & 1.95\% & 2.38\% & 1.60\% & \textbf{0.96\%} \\
    \bottomrule
    \end{tabular}%
  \label{tab:bpp}%
\end{table}%

\subsection{Results with Different LLMs}
\label{app:different_llm}

We further investigate the robustness of our method with respect to different large language model backbones on the Knapsack Problem (KP) under the constructive framework. 
Specifically, we conduct controlled experiments using multiple representative LLMs while keeping all other components and hyperparameters fixed to ensure a fair comparison. 
All results are averaged over three independent runs to mitigate the impact of stochasticity. 
The complete results are summarized in Table~\ref{tab:different_table}.

Overall, the results indicate that AHD methods exhibit consistently stable performance across a wide range of LLM backbones, demonstrating strong model-level generalization.

\begin{table}[htbp]
  \centering
  \caption{Performance comparison of our method with different LLMs backbones on KP under constructive framework.}
    \begin{tabular}{lcccc}
    \toprule
    \multicolumn{1}{c}{Scale} & 50    & 100   & 200   & 500 \\
    \midrule
    DeepSeek-v3 & 20.013 & 40.250 & 57.421 & 90.939 \\
    DeepSeekr1 & 20.003 & 40.238 & 57.405 & 90.918 \\
    GLM-4.7 & 19.998 & 40.236 & 57.405 & 90.924 \\
    Qwen3-30b & 19.996 & 40.236 & 57.399 & 90.918 \\
    Qwen3-max & 20.029 & 40.264 & 57.437 & 90.954 \\
    \bottomrule
    \end{tabular}%
  \label{tab:different_table}%
\end{table}%

\subsection{Results with Complex Optimization Tasks}
\label{app:complex}
Owing to the generality of large language models, our approach can automatically design heuristics for more complex problems and algorithmic frameworks, substantially reducing the reliance on expert knowledge and manual effort. In this section, we provide more detailed experimental results for the DPP under the GA framework and for Mountain Car under RL framework.

\paragraph{DPP under GA Framework}
We evaluate the DPP under the GA framework, which is an iterative evolutionary optimization approach that progressively refines solutions through selection, crossover, and mutation operations. This framework is particularly suitable for complex optimization tasks (e.g., DPP) due to its ability to explore a large and complex search space and to continuously improve solution quality over successive generations.

To provide clearer insights into the optimization behavior during the evolutionary process, we explicitly report the intermediate performance of LLM-based AHD methods. Specifically, Table \ref{tab:dpp} presents the results of ReEvo and TIDE, each executed for three independent runs with a total of 10 evolutionary generations. This evaluation protocol allows us to better observe how solution quality evolves over time and facilitates a transparent comparison of evolutionary dynamics across different methods.

\begin{table}[htbp]
  \centering
  \caption{Evolutionary Performance of ReEvo and TIDE on DPP under GA framework. Results are reported over three runs with 10 evolutionary generations. Generation 0 denotes the initial population before evolution. We show the performance at each generation and the average across runs. Higher values indicate better solution quality. The best average performance is highlighted in bold.}
    \begin{tabular}{clccccccccccc}
    \toprule
    \multicolumn{2}{c}{Generation} & 0     & 1     & 2     & 3     & 4     & 5     & 6     & 7     & 8     & 9     & 10 \\
    \midrule
    \multirow{4}[4]{*}{ReEvo} & run1  & 9.92  & 12.16 & 12.40 & 12.52 & 12.60 & 12.63 & 12.67 & 12.69 & 12.70 & 12.73 & 12.75 \\
          & run2  & 9.93  & 12.20 & 12.38 & 12.52 & 12.60 & 12.64 & 12.68 & 12.70 & 12.72 & 12.73 & 12.74 \\
          & run3  & 9.94  & 12.16 & 12.38 & 12.56 & 12.63 & 12.67 & 12.70 & 12.71 & 12.73 & 12.73 & 12.74 \\
\cmidrule{2-13}          & avg.  & 9.93  & 12.17 & 12.39 & 12.53 & 12.61 & 12.65 & 12.68 & 12.70 & 12.72 & 12.73 & 12.74 \\
    \midrule
    \multirow{4}[4]{*}{TIDE} & run1  & 9.94  & 12.24 & 12.44 & 12.59 & 12.77 & 12.82 & 12.85 & 12.90 & 12.97 & 13.00 & 13.01 \\
          & run2  & 9.90  & 12.11 & 12.32 & 12.52 & 12.58 & 12.64 & 12.67 & 12.70 & 12.72 & 12.73 & 12.75 \\
          & run3  & 9.93  & 12.15 & 12.34 & 12.46 & 12.51 & 12.55 & 12.61 & 12.62 & 12.63 & 12.65 & 12.65 \\
\cmidrule{2-13}          & avg.  & 9.92  & 12.16 & 12.37 & 12.53 & 12.62 & 12.67 & 12.71 & 12.74 & 12.78 & 12.79 & \textbf{12.80} \\
    \bottomrule
    \end{tabular}%
  \label{tab:dpp}%
\end{table}%

\paragraph{Mountain Car under RL Framework}

\begin{table}
  \centering
  \caption{Performance Comparison of LLM-based AHD methods on Mountain Car-v0 under RL framework. Results are averaged over three runs. We report the number of steps required to reach the goal in Mountain Car-v0 for each run and their average. Lower values indicate better policy performance. The best average result is highlighted in bold.}
    \begin{tabular}{ccccc}
    \toprule
    Methods & run1  & run2  & run3  & avg. \\
    \midrule
    EoH   & 118   & 92    & 100   & 103.3 \\
    ReEvo & 85    & 117   & 117   & 106.3 \\
    HSEvo & 114   & 117   & 119   & 116.7 \\
    MCTS-AHD & 124   & 103   & 90    & 105.7 \\
    TIDE  & 85    & 109   & 101   & \textbf{98.3} \\
    \bottomrule
    \end{tabular}%
  \label{tab:car}%
\end{table}%

Mountain Car-v0 is a classic control task in the reinforcement learning setting, where an underpowered car must learn a policy to drive up a steep hill by leveraging momentum through oscillatory movements. The objective is to reach the goal position with as few steps as possible, making the task a standard benchmark for evaluating policy optimization methods.

In this setting, we evaluate our method under the reinforcement learning framework, and Table \ref{tab:car} reports a detailed comparison with existing LLM-based AHD methods, highlighting the effectiveness of the heuristics designed by our approach in policy optimization tasks.

\subsection{Comparative Results}
\label{app:comparative}

In this section, we conduct a comprehensive analysis to validate the internal mechanisms of our framework, specifically focusing on optimization efficiency, evolutionary topology, and adaptive strategy selection. Unless otherwise stated, all analyses presented below are based on the training results of the TSP-Constructive with problem size 50. The experimental configurations and hyperparameter settings are strictly aligned with those used in the main experiments.

\paragraph{Synergy Analysis: Structural Search vs. Parameter Tuning}

\begin{figure}[H]
    \centering
    \includegraphics[width=\linewidth]{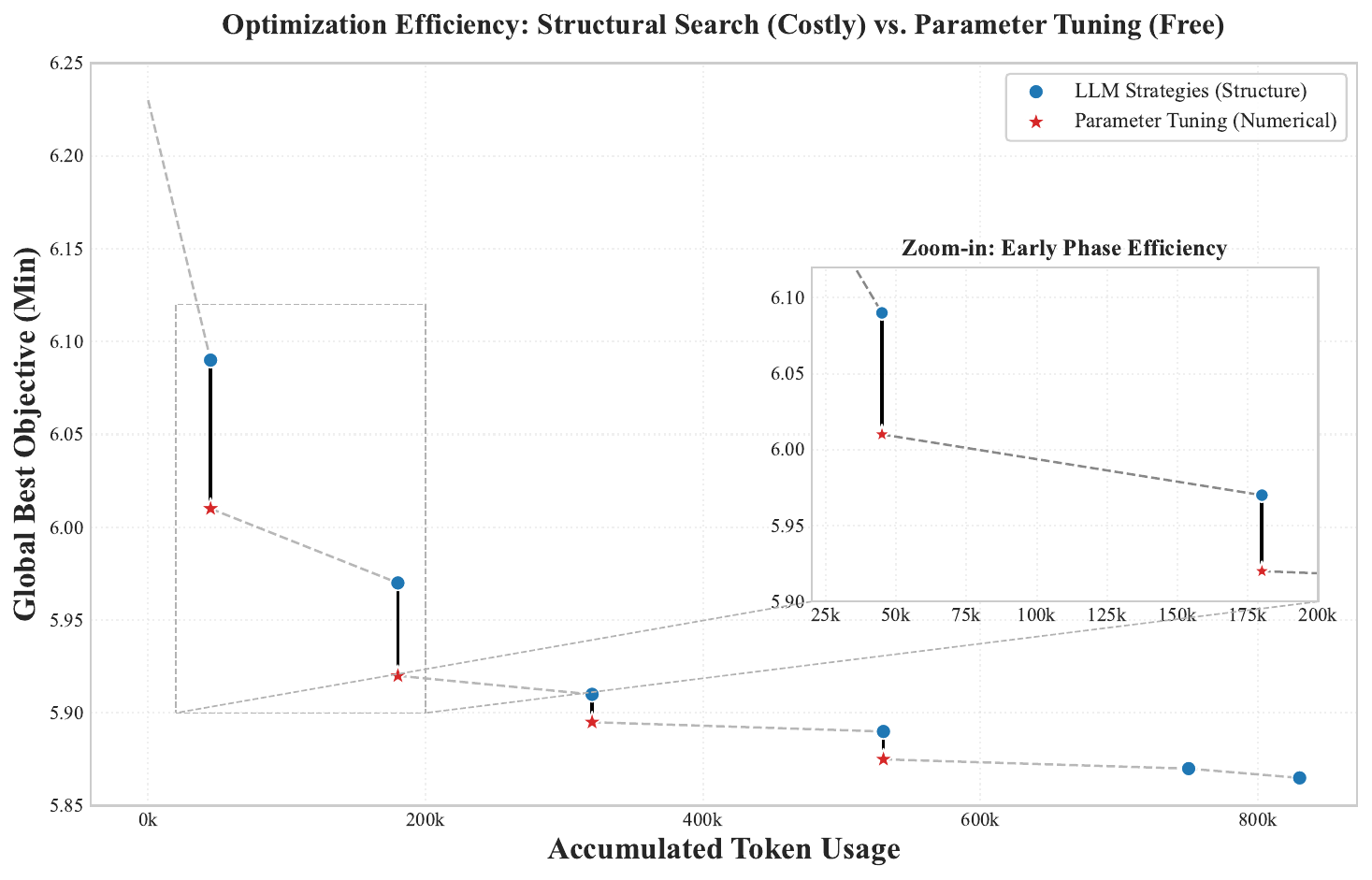}
    \caption{Optimization Efficiency Analysis on TSP-50.}
    \label{fig:efficiency}
\end{figure}

The optimization process visualized in Figure \ref{fig:efficiency} reveals three critical insights regarding the internal dynamics of our framework.
First, the trajectory highlights the cost-efficiency of the parameter tuning module. The blue dots represent the structural search performed by the LLM, which incurs a substantial token cost. In contrast, the parameter tuning module optimizes numerical constants within these structures without consuming tokens. Consequently, the distinct vertical drops (black lines) demonstrate that the tuning module contributes significant performance gains at zero marginal token cost.
Second, the plot evidences a \textbf{synergistic co-evolution} where structural search and parameter tuning mutually reinforce each other. The tuned solutions (red stars) often surpass the subsequent raw LLM-generated candidates, indicating that parameter refinement is essential for realizing the full potential of a code structure. This collaboration leads to substantial performance leaps which would be inefficient to achieve via structural sampling alone.
Third, the overall trajectory exhibits a characteristic staircase pattern. This morphological feature aligns with our algorithmic design, where the LLM conducts exploration in the discrete code space (the horizontal or gentle slopes) while the tuning module performs exploitation in the continuous parameter space (the vertical drops). This pattern confirms that our framework effectively balances global structural exploration with local numerical exploitation.

\paragraph{Evolutionary Topology and Structural Diversity Analysis.}

\begin{figure}
    \centering
    \includegraphics[width=\linewidth]{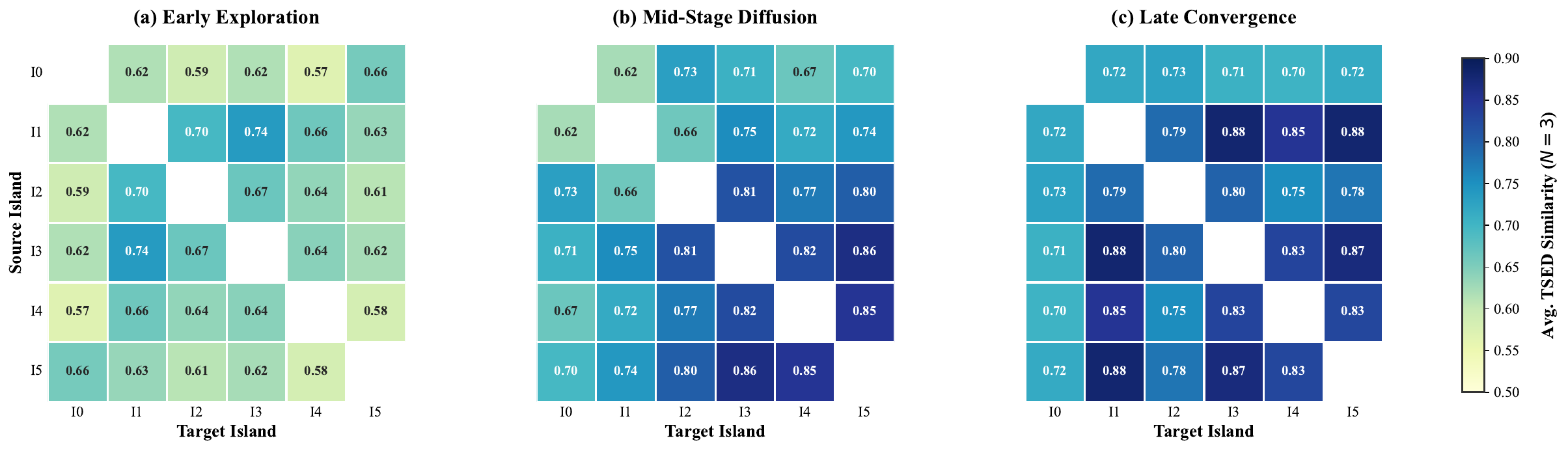}
    \caption{Evolution of Structural Similarity (TSED) across islands.}
    \label{fig:tsed_evolution}
\end{figure}

Figure \ref{fig:tsed_evolution} visualizes the structural dynamics of the population through three evolutionary phases.

First, during the \textit{Early Exploration} phase (Panel a), similarity scores remain uniformly low (avg $\approx$ 0.62). Since these values fall below the structural threshold (0.7), the system predominantly triggers \textit{Insight Transfer}. This mechanism allows islands to share high-level design rationales derived from best-worst comparisons without overwriting the actual code structure. Consequently, the heatmap shows that islands improve while maintaining distinct algorithmic trajectories (light colors), effectively preserving population diversity.

Second, the \textit{Mid-Stage Diffusion} phase (Panel b) captures the transition point. As certain islands (I3, I4, I5) converge on similar optimal structures, their similarity crosses the threshold, triggering \textit{Code Transfer}. This switches the mode to direct exploitation, leading to the rapid formation of a high-similarity cluster (deep blue blocks). In contrast, Island 0 remains structurally distinct (similarity $<$ 0.7); therefore, it continues to receive only high-level insights. This prevents forced homogenization and allows Island 0 to persist as a unique structural niche.

Third, the \textit{Late Convergence} phase (Panel c) demonstrates the long-term stability of this dual mechanism. While the dominant cluster solidifies via Code Transfer, the persistence of Island 0 (similarity $\approx$ 0.71) confirms that Insight Transfer successfully prevents total mode collapse. This proves that the adaptive switching mechanism effectively balances the rapid exploitation of successful structures with the preservation of alternative algorithmic logic.

\begin{figure}
    \centering
    \includegraphics[width=\linewidth]{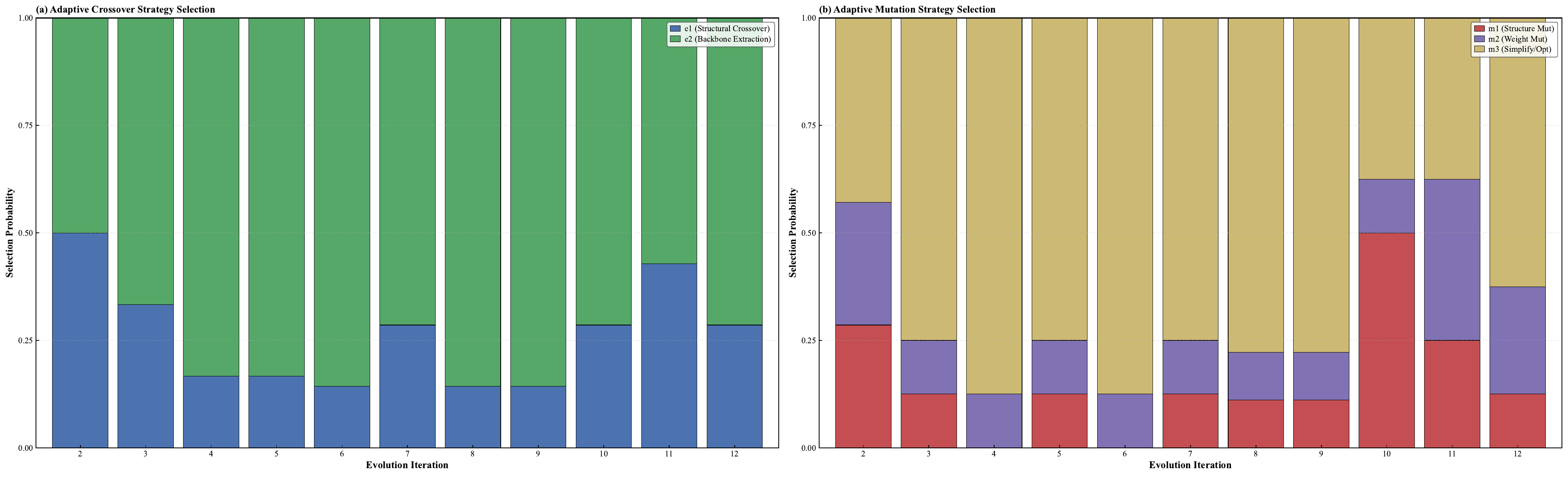}
    \caption{Dynamics of Adaptive Strategy Selection via UCB}
    \label{fig:ucb_distribution}
\end{figure}

\paragraph{Analysis of Adaptive Strategy Selection}
We analyzed the temporal distribution of operator selection probabilities. Figure \ref{fig:ucb_distribution} reveals that the system spontaneously evolves a multi-stage optimization strategy.
In the Crossover phase (Panel a), the algorithm maintains a balanced usage of Structural Crossover (e1) and Backbone Extraction (e2) in the initial iterations. However, as the population matures, e2 gradually becomes the dominant strategy. This trend indicates that the system learns to shift from aggressively mixing diverse structures (Exploration) to consolidating and exploiting successful algorithmic backbones (Exploitation) to stabilize convergence.

More notably, the Mutation phase (Panel b) exhibits a distinct \textit{clean-then-refine curriculum}. The Simplification operator (m3, yellow bars) dominates the early-to-mid stages. As evidenced by the substantial complexity of the evolved TSP heuristic presented in \textbf{Appendix ~\ref{sec:tsp_con}}, LLM-generated solutions for hard combinatorial problems often initiate with verbose control flows and redundant logic. The UCB selector effectively identifies this structural bloat as the primary bottleneck, prioritizing simplification to distill the algorithmic backbone. Subsequently, in the late stages (e.g., Iterations 10-12), we observe a significant resurgence of Structural (m1) and Weight Mutation (m2). This transition confirms that once the code logic is streamlined, the system adaptively switches focus to fine-grained parameter tuning and structural perturbation to escape local optima.

\subsection{Consumption of Time and Token}
We further analyze the computational cost of different LLM-based heuristic design methods in terms of time and token consumption. Table~\ref{tab:cost} reports the time cost as well as the total numbers of input and output tokens consumed across different problem settings and algorithmic frameworks. 
All statistics are collected under the same experimental setup to ensure a fair comparison. All methods are configured according to the settings described in the original papers.

As shown in the Table \ref{tab:cost}, different methods exhibit distinct cost profiles depending on their search strategies and interaction patterns with large language models.

\begin{table}[h]
  \centering
  \caption{Comparison of time cost and token consumption for different LLM-based AHD methods.}
    \begin{tabular}{c|c|cccc}
    \toprule
    Methods & Consumption & TSP-Constructive & KP-Constructive & TSP-ACO & MKP-ACO \\
    \midrule
    \multirow{3}[2]{*}{EoH} & Time  & 3h    & 5h    & 4h    & 4h \\
          & Input Token & 0.5M  & 0.6M  & 0.4M  & 0.5M \\
          & Output Token & 0.2M  & 0.2M  & 0.2M  & 0.2M \\
    \midrule
    \multirow{3}[2]{*}{ReEvo} & Time  & 1.5h  & 2h    & 0.6h  & 3h \\
          & Input Token & 0.6M  & 0.6M  & 0.2M  & 0.8M \\
          & Output Token & 0.2M  & 0.2M  & 0.1M  & 0.2M \\
    \midrule
    \multirow{3}[2]{*}{HSEvo} & Time  & 5h  & 2.5h  & 2h    & 3h \\
          & Input Token & 1.7M  & 0.6M  & 0.8M  & 0.6M \\
          & Output Token & 0.6M  & 0.2M  & 0.2M  & 0.2M \\
    \midrule
    \multirow{3}[2]{*}{MCTS-AHD} & Time  & 8.5h  & 5h    & 8h    & 7h \\
          & Input Token & 1.5M  & 1.1M  & 1.2M  & 1.5M \\
          & Output Token & 0.6M  & 0.4M  & 0.5M  & 0.5M \\
    \midrule
    \multirow{3}[2]{*}{TIDE} & Time  & 8h    & 2.5h  & 3h    & 4h \\
          & Input Token & 0.8M  & 0.7M  & 1M    & 0.6M \\
          & Output Token & 0.4M  & 0.4M  & 0.5M  & 0.4M \\
    \bottomrule
    \end{tabular}%
  \label{tab:cost}%
\end{table}%

It is worth noting that ReEvo, HSEvo, and TIDE adopt a warm-start strategy, where seed functions are used to initialize the population. These methods aim to expand and refine existing heuristics to escape local optima. Their effectiveness crucially depends on how the initial seed functions are exploited and extended during the search process.

In the original papers of ReEvo and HSEvo, they adopt fixed evaluation budgets (e.g., 100 or 450 evaluations), under which the methods should have converged.  

Due to the relatively small evaluation budgets adopted in their default settings, ReEvo and HSEvo exhibit lower computational costs. To more fairly assess the potential of warm-start methods and examine whether increased computational budgets lead to more substantial performance improvements, we further evaluate these methods under a unified budget of 800 evaluations, and compare the resulting solution quality as well as the corresponding computational cost in Table~\ref{tab:additional_cost}. 

To better characterize the effectiveness and efficiency of seed-based heuristic improvement, we introduce one additional evaluation metric. The metric normalizes this improvement by the total number of tokens consumed, measuring the relative performance gain per token with respect to the seed heuristic. It captures cost-aware efficiency, quantifying how effectively computational resources are converted into heuristic improvements. In the table, \textit{Improve} denotes the relative improvement of the designed heuristics over the seed functions per token (in millions, M), which is computed as the relative gap (in \%) between the designed heuristics and the seed functions divided by the total number of tokens consumed during evolution.

\begin{table}[htbp]
  \centering
  \caption{Comparison of cost and improvement efficiency for ReEvo, HSEvo and TIDE on KP constructive and TSP ACO. \textit{Gap} represents the result of heuristics after 800 evaluations relative to the optimal solution (optimal solutions for KP are obtained using OR-Tools, and for TSP using LKH3~\citep{lin1973effective}, and lower is better). \textit{Improve} denotes the relative improvement (\%) over seed functions per M tokens (larger is better).}
    \begin{tabular}{c|c|ccc|c|cc|cc}
    \toprule
    \multicolumn{10}{c}{KP-Constructive} \\
    \midrule
    \multirow{2}[2]{*}{Methods} & \multirow{2}[2]{*}{Time} & \multicolumn{3}{c|}{ Tokens} & \multirow{2}[2]{*}{LLM requests} & \multicolumn{2}{c|}{100} & \multicolumn{2}{c}{200} \\
          &       & Input & Output & Total &       & Gap   & Improve & Gap   & Improve \\
    \midrule
    ReEvo & 2.5h  & 1.2M  & 0.3M  & 1.5M  & 1382  & 0.08\% & 0.023\% & 0.07\% & 0.013\% \\
    HSEvo & 5h    & 0.7M  & 0.2M  & 0.9M  & 589   & 0.11\% & 0.012\% & 0.08\% & 0.008\% \\
    TIDE  & 2.5h  & 0.7M  & 0.4M  & 1.1M  & 788   & 0.02\% & 0.089\% & 0.02\% & 0.066\% \\
    \midrule
    \multicolumn{10}{c}{TSP-ACO} \\
    \midrule
    \multirow{2}[2]{*}{Methods} & \multirow{2}[2]{*}{Time} & \multicolumn{3}{c|}{ Tokens} & \multirow{2}[2]{*}{LLM requests} & \multicolumn{2}{c|}{50} & \multicolumn{2}{c}{100} \\
          &       & Input & Output & Total &       & Gap   & Improve & Gap   & Improve \\
    \midrule
    ReEvo & 3.5h  & 1.5M  & 0.3M  & 1.8M  & 1384  & 2.15\% & 2.69\% & 6.16\% & 6.27\% \\
    HSEvo & 4h    & 1.2M  & 0.3M  & 1.5M  & 589   & 1.72\% & 3.50\% & 5.25\% & 8.03\% \\
    TIDE  & 3h    & 1M    & 0.5M  & 1.5M  & 788   & 1.60\% & 3.57\% & 4.37\% & 8.52\% \\
    \bottomrule
    \end{tabular}%
  \label{tab:additional_cost}%
\end{table}%

\section{Prompt Engineering Details.}

In this section, we provide the comprehensive prompt templates utilized in our evolutionary framework. To ensure reproducibility and maintain methodological consistency with state-of-the-art approaches, our prompt engineering strategy builds upon the foundational operators proposed in the EOH~\citep{liu2024evolution}. Specifically, the recombination strategies (E1, E2) and mutation strategies (M1, M3) are adopted from EoH. The M2 operator has been fine-tuned to emphasize \textit{scoring logic refinement}, directing the LLM to analyze decision boundaries rather than performing generic parameter perturbation.

Distinctively, to support the specific dynamics of our distributed island model, we introduce two novel prompt mechanisms: \textit{Meta-Cognitive Insight Extraction} and \textit{Selective Reset}. These additions are critical for facilitating high-level knowledge transfer between sub-populations and enabling knowledge-guided recovery from deep local optima.

\subsection{Recombination Prompt Strategy (E1)}
The E1 operator serves as an exploration-centric crossover mechanism. Unlike standard genetic crossover that splices gene sequences, E1 leverages the LLM to synthesize a structurally novel algorithm by analyzing $k$ parent solutions selected via tournament. Crucially, this prompt integrates \textit{meta-cognitive insights}, defined as the accumulated knowledge from the island's evolutionary history, to guide the generation process away from known failures and towards promising heuristics. Furthermore, the prompt structure ensures valid Python execution and enforces a Chain-of-Thought (CoT) process via the \textit{[Thought]} and \textit{[KEY PARAMETERS]} blocks.
\begin{tcolorbox}[
  enhanced, breakable, colback=white, colframe=navy, boxrule=1.5pt, arc=4mm,
  top=4mm, bottom=4mm, left=4mm, right=4mm,
  colbacktitle=navy, coltitle=white, fonttitle=\large\bfseries,
  title=Prompt Template for Operator E1 (Explorative Crossover)
]
\small
\textbf{[System Message]} \\
You are an expert algorithm engineer. Design efficient heuristic functions. Output your algorithm description inside a brace, then implement it in Python.

\vspace{0.2cm}
\hrule
\vspace{0.2cm}

\textbf{[User Message]} \\
\textbf{Problem Description:} \textcolor{mygreen}{\{\{self.problem\_desc\}\}} \\
\textbf{Function Signature:} \textcolor{blue}{\{\{self.func\_name\}\}} \\
\textbf{Goal:} Design a novel heuristic algorithm.

I have \textcolor{red}{$k$} existing algorithms: \\
\textbf{No.1 algorithm:} \\
\textit{\{\{Parent 1 Thought\}\}} \\
Code:
\begin{verbatim}
{{Parent 1 Code}}
\end{verbatim}
\dots \\
\textbf{No.$k$ algorithm:} \\
\dots

\textbf{Task:} Implement as a function named \textcolor{blue}{\{\{self.func\_name\}\}\_v2} with the same signature.
Create a new algorithm that has a \textbf{totally different form} from the given ones.

You must strictly follow this output format:
\begin{itemize}[leftmargin=*, noitemsep, topsep=0pt]
    \item \texttt{[Thought]}: Summarize in exactly 2 sentences: the core idea.
    \item \texttt{[KEY PARAMETERS]}: List a moderate number of tunable parameters and their roles (avoid too many hardcoded values).
    \item \texttt{[Code]}: Complete executable Python code in a code block. Include all imports.
\end{itemize}

\vspace{0.2cm}
\textit{\textbf{Contextual Insight:}} \\
\textcolor{purple}{\{\{Local Island Insight\}\}} \\
\textcolor{purple}{\{\{Neighbor Island Insight (if applicable)\}\}}
\end{tcolorbox}

\subsection{Recombination Prompt Strategy (E2)}
The E2 operator functions as a \textit{Conceptual Crossover} mechanism. While standard crossover exchanges code fragments, E2 instructs the LLM to abstract the underlying logic shared by the parent solutions. By explicitly requesting the identification of a \textit{common backbone,} this operator synthesizes a new heuristic that retains the structural consensus of high-performing parents while introducing novel variations in implementation. This approach effectively stabilizes beneficial traits while preventing the population from collapsing into identical code clones.

\begin{tcolorbox}[
  enhanced, breakable, colback=white, colframe=navy, boxrule=1.5pt, arc=4mm,
  top=4mm, bottom=4mm, left=4mm, right=4mm,
  colbacktitle=navy, coltitle=white, fonttitle=\large\bfseries,
  title=Prompt Template for Operator E2 (Backbone Extraction)
]
\small
\textbf{[System Message]} \\
You are an expert algorithm engineer. Design efficient heuristic functions. Output your algorithm description inside a brace, then implement in Python.

\vspace{0.2cm}
\hrule
\vspace{0.2cm}

\textbf{[User Message]} \\
\textbf{Problem Description:} \textcolor{mygreen}{\{\{self.problem\_desc\}\}} \\
\textbf{Function Signature:} \textcolor{blue}{\{\{self.func\_name\}\}} \\
\textbf{Goal:} Design a novel heuristic algorithm.

I have \textcolor{red}{$k$} existing algorithms: \\
\textbf{No.1 algorithm:} \\
\textit{\{\{Parent 1 Thought\}\}} \\
Code:
\begin{verbatim}
{{Parent 1 Code}}
\end{verbatim}
\dots \\
\textbf{No.$k$ algorithm:} \\
\dots

\textbf{Task:} Implement as a function named \textcolor{blue}{\{\{self.func\_name\}\}\_v2} with the same signature.
Identify the \textbf{common backbone idea} in these algorithms, then create a new algorithm motivated from it but with a different form.

You must strictly follow this output format:
\begin{itemize}[leftmargin=*, noitemsep, topsep=0pt]
    \item \texttt{[Thought]}: Summarize in exactly 2 sentences: the core idea.
    \item \texttt{[KEY PARAMETERS]}: List a moderate number of tunable parameters and their roles (avoid too many hardcoded values).
    \item \texttt{[Code]}: Complete executable Python code in a code block. Include all imports.
\end{itemize}

\vspace{0.2cm}
\textit{\textbf{Contextual Insight:}} \\
\textcolor{purple}{\{\{Local Island Insight\}\}} \\
\textcolor{purple}{\{\{Neighbor Island Insight (if applicable)\}\}}
\end{tcolorbox}

\subsection{Mutation Prompt Strategy (M1)}
The M1 operator implements a \textit{Structural Mutation} mechanism. Unlike simple parameter perturbation, M1 prompts the LLM to generate a variant that possesses a different algorithmic form while remaining a modified version of the parent. A distinguishing feature of this operator is the injection of cross-island knowledge: the prompt context optionally includes insights from a neighboring island. This allows the mutation process to be guided not only by local history but also by external success strategies, effectively bridging the gap between local exploitation and global exploration.

\begin{tcolorbox}[
  enhanced, breakable, colback=white, colframe=navy, boxrule=1.5pt, arc=4mm,
  top=4mm, bottom=4mm, left=4mm, right=4mm,
  colbacktitle=navy, coltitle=white, fonttitle=\large\bfseries,
  title=Prompt Template for Operator M1 (Structural Mutation)
]
\small
\textbf{[System Message]} \\
You are an expert algorithm engineer. Design efficient heuristic functions. Output your algorithm description inside a brace, then implement in Python.

\vspace{0.2cm}
\hrule
\vspace{0.2cm}

\textbf{[User Message]} \\
\textbf{Problem Description:} \textcolor{mygreen}{\{\{self.problem\_desc\}\}} \\
\textbf{Function Signature:} \textcolor{blue}{\{\{self.func\_name\}\}} \\
\textbf{Goal:} Design a novel heuristic algorithm.

\textbf{Current algorithm:} \\
\textit{\{\{Parent Thought\}\}} \\
Code:
\begin{verbatim}
{{Parent Code}}
\end{verbatim}

\textbf{Task:} Implement as a function named \textcolor{blue}{\{\{self.func\_name\}\}\_v2} with the same signature.
Create a new algorithm that has a \textbf{different form} but can be a modified version of the provided one.

You must strictly follow this output format:
\begin{itemize}[leftmargin=*, noitemsep, topsep=0pt]
    \item \texttt{[Thought]}: Summarize in exactly 2 sentences: the core idea.
    \item \texttt{[KEY PARAMETERS]}: List a moderate number of tunable parameters and their roles.
    \item \texttt{[Code]}: Complete executable Python code in a code block. Include all imports.
\end{itemize}

\vspace{0.2cm}
\textit{\textbf{Contextual Insight:}} \\
\textcolor{purple}{\{\{Local Island Insight\}\}} \\
\textcolor{purple}{\{\{Neighbor Island Insight (Injection from adjacent population)\}\}}
\end{tcolorbox}

\subsection{Mutation Prompt Strategy (M2)}
The M2 operator implements a \textit{Scoring Logic Refinement} strategy. While M1 focuses on high-level architectural changes, M2 directs the LLM to analyze the parent's internal decision-making process, specifically targeting the \textit{scoring function} and its constituent weightings. This operator instructs the model to deconstruct the existing scoring mechanics and synthesize a variant with modulated behaviors or alternative logic paths. By integrating contextual insights, M2 enables the evolutionary process to perform nuanced structural adjustments that refine how the heuristic prioritizes conflicting objectives, thereby sharpening the solution’s precision without discarding its proven backbone.

\begin{tcolorbox}[
  enhanced, breakable, colback=white, colframe=navy, boxrule=1.5pt, arc=4mm,
  top=4mm, bottom=4mm, left=4mm, right=4mm,
  colbacktitle=navy, coltitle=white, fonttitle=\large\bfseries,
  title=Prompt Template for Operator M2 (Scoring Logic Refinement)
]
\small
\textbf{[System Message]} \\
You are an expert algorithm engineer. Design efficient heuristic functions. Output your algorithm description inside a brace, then implement in Python.

\vspace{0.2cm}
\hrule
\vspace{0.2cm}

\textbf{[User Message]} \\
\textbf{Problem Description:} \textcolor{mygreen}{\{\{self.problem\_desc\}\}} \\
\textbf{Function Signature:} \textcolor{blue}{\{\{self.func\_name\}\}} \\
\textbf{Goal:} Design a novel heuristic algorithm.

\textbf{Current algorithm:} \\
\textit{\{\{Parent Thought\}\}} \\
Code:
\begin{verbatim}
{{Parent Code}}
\end{verbatim}

\textbf{Task:} Implement as a function named \textcolor{blue}{\{\{self.func\_name\}\}\_v2} with the same signature.
Identify the \textbf{main scoring components} and create a new algorithm with different \textbf{configurations or score function}.

You must strictly follow this output format:
\begin{itemize}[leftmargin=*, noitemsep, topsep=0pt]
    \item \texttt{[Thought]}: Summarize in exactly 2 sentences: the core idea.
    \item \texttt{[KEY PARAMETERS]}: List the key controllable factors and their roles (for downstream optimization).
    \item \texttt{[Code]}: Complete executable Python code in a code block. Include all imports.
\end{itemize}

\vspace{0.2cm}
\textit{\textbf{Contextual Insight:}} \\
\textcolor{purple}{\{\{Local Island Insight\}\}}
\end{tcolorbox}

\subsection{Mutation Prompt Strategy (M3)}
The M3 operator functions as a Regularization Mutation strategy designed to enhance heuristic robustness and combat overfitting. It instructs the LLM to perform a critical analysis of the parent function, specifically targeting components suspected of being over-specialized to in-distribution data. These potentially brittle or overly complex segments are then strategically pruned or streamlined. The outcome is a more parsimonious and computationally lean implementation theorized to exhibit superior generalization to out-of-distribution scenarios while preserving the original function signature to ensure architectural compatibility.

\begin{tcolorbox}[
  enhanced, breakable, colback=white, colframe=navy, boxrule=1.5pt, arc=4mm,
  top=4mm, bottom=4mm, left=4mm, right=4mm,
  colbacktitle=navy, coltitle=white, fonttitle=\large\bfseries,
  title=Prompt Template for Operator M3 (Regularization Mutation)
]
\small
\textbf{[System Message]} \\
You are an expert algorithm engineer. Design efficient heuristic functions. Output your algorithm description inside a brace, then implement in Python.

\vspace{0.2cm}
\hrule
\vspace{0.2cm}

\textbf{[User Message]} \\
\textbf{Problem Description:} \textcolor{mygreen}{\{\{self.problem\_desc\}\}} \\
\textbf{Function Signature:} \textcolor{blue}{\{\{self.func\_name\}\}} \\
\textbf{Goal:} Design a novel heuristic algorithm.

\textbf{Current algorithm Code:}
\begin{verbatim}
{{Parent Code}}
\end{verbatim}

\textbf{Task:} Implement as a function named \textcolor{blue}{\{\{self.func\_name\}\}\_v2} with the same signature.

Identify the \textbf{main components} in the function below. Analyze whether any components can be \textbf{overfit} to specific instances. \textbf{Simplify or optimize} the components to enhance generalization. Provide the different revised code, keeping the function name, inputs, and outputs unchanged.

You must strictly follow this output format:
\begin{itemize}[leftmargin=*, noitemsep, topsep=0pt]
    \item \texttt{[Thought]}: Summarize in exactly 2 sentences: the core idea.
    \item \texttt{[KEY PARAMETERS]}: List a moderate number of tunable parameters and their roles.
    \item \texttt{[Code]}: Complete executable Python code in a code block. Include all imports.
\end{itemize}

\vspace{0.2cm}
\textit{\textbf{Contextual Insight:}} \\
\textcolor{purple}{\{\{Local Island Insight\}\}}
\end{tcolorbox}

\subsection{Meta-Cognitive Prompt Strategy (Insight Extraction)}
To prevent the evolutionary search from becoming a blind trial-and-error process, the framework employs a dedicated Insight Extraction mechanism. This operator functions as a meta-cognitive feedback loop that converts raw code execution data into semantic design principles. Rather than analyzing a solution in isolation, the system employs a \textit{contrastive prompting strategy}: it presents the LLM with the island's current Elite (best performer) alongside the local Straggler (worst performer). By forcing the model to explicitly articulate the mechanistic reasons behind the performance gap, the system distills abstract strategies from concrete implementations. These generated insights are then stored in the island's memory and re-injected into subsequent mutation and crossover prompts, thereby guiding future generations with accumulated architectural knowledge.

\begin{tcolorbox}[
  enhanced, breakable, colback=white, colframe=navy, boxrule=1.5pt, arc=4mm,
  top=4mm, bottom=4mm, left=4mm, right=4mm,
  colbacktitle=navy, coltitle=white, fonttitle=\large\bfseries,
  title=Prompt Template for Insight Extraction (Contrastive Analysis)
]
\small
\textbf{[System Message]} \\
You are an expert in the domain of optimization heuristics. Identify the mechanisms responsible for the performance gap and explain why it is effective in two paragraphs.

\vspace{0.2cm}
\hrule
\vspace{0.2cm}

\textbf{[User Message]} \\
\textbf{High Performance (Score: \textcolor{mygreen}{\{\{Best Objective\}\}}):}
\begin{verbatim}
{{Best Code}}
\end{verbatim}

\vspace{0.2cm}
\textbf{Low Performance (Score: \textcolor{red}{\{\{Worst Objective\}\}}):}
\begin{verbatim}
{{Worst Code}}
\end{verbatim}

\textbf{Instruction:} \\
You are an expert in the domain of optimization heuristics. Identify the mechanisms responsible for the performance gap and explain why it is effective in two paragraphs.
\end{tcolorbox}

\subsection{Selective Reset Prompt Strategy (Knowledge-Guided Restart)}
To mitigate the risk of irreversible convergence where an island population becomes trapped in a deep local optimum, the framework employs a Selective Reset mechanism. Unlike traditional restart strategies that re-initialize populations randomly, effectively discarding all accumulated optimization progress, this operator leverages the \textit{Global Elite heuristic} to perform a \textit{knowledge-guided restart}. When deep stagnation is detected via trajectory analysis, the system retrieves a high-performing solution from the global archive. The LLM is then prompted to extract the underlying domain insights from this elite reference and synthesize a \textit{structurally novel} algorithm derived from it. This process effectively transplants the global state-of-the-art logic into the stagnant island but forces a divergent implementation path, thereby reinvigorating the local search space with high-potential genetic material.

\begin{tcolorbox}[
  enhanced, breakable, colback=white, colframe=navy, boxrule=1.5pt, arc=4mm,
  top=4mm, bottom=4mm, left=4mm, right=4mm,
  colbacktitle=navy, coltitle=white, fonttitle=\large\bfseries,
  title=Prompt Template for Selective Reset (Knowledge-Guided Restart)
]
\small
\textbf{[System Message]} \\
You are an expert algorithm engineer. Design efficient heuristic functions. Output your algorithm description inside a brace, then implement in Python.

\vspace{0.2cm}
\hrule
\vspace{0.2cm}

\textbf{[User Message]} \\
\textbf{Reference Elite (Score: \textcolor{mygreen}{\{\{Global Elite Score\}\}}):}
\begin{verbatim}
{{Global Elite Code}}
\end{verbatim}

\textbf{Instruction:} \\
You are an expert in the domain of optimization heuristics. Extract domain insights from the Elite solution provided above, then create a \textbf{structurally novel and advanced algorithm}. Output the design rationale followed by the implementation of \textcolor{blue}{\{\{self.func\_name\}\}\_v2}.

You must strictly follow this output format:
\begin{itemize}[leftmargin=*, noitemsep, topsep=0pt]
    \item \texttt{[Thought]}: Summarize in exactly 2 sentences: the core idea.
    \item \texttt{[KEY PARAMETERS]}: List a moderate number of tunable parameters and their roles.
    \item \texttt{[Code]}: Complete executable Python code in a code block. Include all imports.
\end{itemize}
\end{tcolorbox}

\section{Examples of LLM-Generated Heuristics}

\definecolor{codegreen}{rgb}{0,0.6,0}
\definecolor{codegray}{rgb}{0.5,0.5,0.5}
\definecolor{codepurple}{rgb}{0.58,0,0.82}
\definecolor{backcolour}{rgb}{0.96,0.96,0.96} 

\lstdefinestyle{iclrstyle}{
    backgroundcolor=\color{backcolour},   
    commentstyle=\color{codegreen},
    keywordstyle=\color{magenta}\bfseries, 
    numberstyle=\tiny\color{codegray},
    stringstyle=\color{codepurple},
    basicstyle=\ttfamily\footnotesize,     
    breakatwhitespace=false,         
    breaklines=true,                       
    captionpos=t,                          
    keepspaces=true,                 
    numbers=none,                          
    numbersep=5pt,                  
    showspaces=false,                
    showstringspaces=false,
    showtabs=false,                  
    tabsize=4,
    frame=single,                          
    frameround=tttt,                       
    rulecolor=\color{black!30},            
    xleftmargin=1em,                     
}

\lstset{style=iclrstyle}

\subsection{Best-Performing Heuristic for Constructive TSP}
\label{sec:tsp_con}
\begin{lstlisting}[language=Python]
import heapq
import math

def select_next_node_v2(current_node, destination_node, unvisited_nodes, distance_matrix):
    if not unvisited_nodes:
        return None
    if len(unvisited_nodes) == 1:
        return next(iter(unvisited_nodes))
    
    # Tunable parameters
    alpha = 0.8
    beta = 0.1
    gamma = 0.6
    theta = 0.05
    candidate_ratio = 2.5
    max_candidates = 12

    unvisited_list = list(unvisited_nodes)
    n_total = len(unvisited_list)
    k_candidates = min(int(n_total * candidate_ratio), max_candidates, n_total)
    if k_candidates == 0:
        k_candidates = 1
    candidate_set = heapq.nsmallest(k_candidates, unvisited_list, key=lambda x: distance_matrix[current_node][x])

    def calculate_mst_cost(nodes):
        if len(nodes) <= 1:
            return 0
        nodes = list(nodes)
        edges = []
        for i in range(len(nodes)):
            for j in range(i + 1, len(nodes)):
                u, v = nodes[i], nodes[j]
                edges.append((distance_matrix[u][v], u, v))
        edges.sort()

        parent = {node: node for node in nodes}
        rank = {node: 0 for node in nodes}

        def find(x):
            while parent[x] != x:
                parent[x] = parent[parent[x]]
                x = parent[x]
            return x

        def union(x, y):
            rx, ry = find(x), find(y)
            if rx == ry:
                return False
            if rank[rx] < rank[ry]:
                parent[rx] = ry
            elif rank[rx] > rank[ry]:
                parent[ry] = rx
            else:
                parent[ry] = rx
                rank[rx] += 1
            return True

        mst_cost = 0
        count = 0
        for cost, u, v in edges:
            if union(u, v):
                mst_cost += cost
                count += 1
                if count == len(nodes) - 1:
                    break
        return mst_cost

    def simulate_refined_path(start, remaining_nodes, dest):
        # Construct greedy path: start -> all remaining -> dest
        path = [start]
        current = start
        remaining = set(remaining_nodes)
        if start in remaining:
            remaining.remove(start)
        
        while remaining:
            next_n = min(remaining, key=lambda x: distance_matrix[current][x])
            path.append(next_n)
            current = next_n
            remaining.remove(current)
        path.append(dest)

        # Apply iterative 2-opt until no improvement
        improved = True
        while improved:
            improved = False
            best_gain = 0
            best_i, best_j = -1, -1
            n = len(path)
            for i in range(1, n - 2):
                for j in range(i + 2, n):
                    old_dist = distance_matrix[path[i-1]][path[i]] + distance_matrix[path[j-1]][path[j]]
                    new_dist = distance_matrix[path[i-1]][path[j-1]] + distance_matrix[path[i]][path[j]]
                    gain = old_dist - new_dist
                    if gain > best_gain:
                        best_gain = gain
                        best_i, best_j = i, j
            if best_gain > 1e-9:
                path[best_i:best_j] = reversed(path[best_i:best_j])
                improved = True

        total_cost = sum(distance_matrix[path[i]][path[i+1]] for i in range(len(path)-1))
        return total_cost

    best_score = float('inf')
    next_node = candidate_set[0]

    for candidate in candidate_set:
        remaining = set(unvisited_nodes)
        remaining.discard(candidate)

        direct_cost = distance_matrix[current_node][candidate]
        mst_cost = calculate_mst_cost(remaining) if remaining else 0
        sim_cost = simulate_refined_path(candidate, remaining, destination_node)
        dest_bias = distance_matrix[candidate][destination_node]

        score = (
            alpha * direct_cost +
            beta * mst_cost +
            gamma * sim_cost +
            theta * dest_bias
        )

        if score < best_score:
            best_score = score
            next_node = candidate

    return next_node
\end{lstlisting}

\subsection{Best-Performing Heuristic for Online BPP}
\begin{lstlisting}[language=Python]
import numpy as np

# Global state variables for priority_v2
_min_item_estimate_v2 = None
_recent_bin_usage_v2 = None

def priority_v2(item, bins_remain_cap):
    global _min_item_estimate_v2, _recent_bin_usage_v2

    epsilon = 1e-8
    bins_remain_cap = np.asarray(bins_remain_cap, dtype=float)
    feasible = bins_remain_cap >= item
    if not np.any(feasible):
        return -np.inf * np.ones_like(bins_remain_cap)

    leftover = bins_remain_cap - item

    # Key parameters (streamlined and rebalanced)
    fit_weight = 1.2
    reuse_bonus_weight = 0.8
    min_item_decay = 0.1
    history_bonus_weight = 0.15
    utility_target_ratio = 0.4
    utility_width = 0.25

    # --- Adaptive minimum item estimation ---
    if _min_item_estimate_v2 is None:
        _min_item_estimate_v2 = float(item)
    else:
        _min_item_estimate_v2 = (1 - min_item_decay) * min(_min_item_estimate_v2, item) + min_item_decay * item

    # --- Best-fit score ---
    fit_score = fit_weight / (leftover + epsilon)

    # --- Waste penalty: penalize unusable leftovers (< estimated min item) ---
    usable_threshold = _min_item_estimate_v2
    waste_penalty = np.where(
        leftover < usable_threshold,
        (usable_threshold - leftover) / (usable_threshold + epsilon),
        0.0
    )

    # --- Reuse bonus: Gaussian centered at utility target ---
    estimated_bin_capacity = np.max(bins_remain_cap) if len(bins_remain_cap) > 0 else max(item, 1.0)
    estimated_bin_capacity = max(estimated_bin_capacity, item)
    norm_leftover = leftover / (estimated_bin_capacity + epsilon)
    deviation = (norm_leftover - utility_target_ratio) / (utility_width + epsilon)
    reuse_bonus = reuse_bonus_weight * np.exp(-0.5 * deviation ** 2)

    # --- History bonus: favor recently used bins ---
    history_bonus = np.zeros_like(bins_remain_cap)
    if _recent_bin_usage_v2 is not None:
        for idx, count in _recent_bin_usage_v2.items():
            if idx < len(history_bonus):
                history_bonus[idx] += history_bonus_weight * count

    # Compute raw priorities to select bin
    raw_priorities = fit_score + reuse_bonus - waste_penalty
    raw_priorities = np.where(feasible, raw_priorities, -np.inf)
    chosen_bin_idx = int(np.argmax(raw_priorities))

    # Update usage history with decay
    if _recent_bin_usage_v2 is None:
        _recent_bin_usage_v2 = {}
    _recent_bin_usage_v2 = {k: v * 0.9 for k, v in _recent_bin_usage_v2.items() if v * 0.9 > 0.01}
    _recent_bin_usage_v2[chosen_bin_idx] = _recent_bin_usage_v2.get(chosen_bin_idx, 0) + 1.0

    # Final priority including history
    priorities = fit_score + reuse_bonus - waste_penalty + history_bonus
    priorities = np.where(feasible, priorities, -np.inf)

    return priorities
\end{lstlisting}

\subsection{Best-Performing Heuristic for Offline BPP-ACO}
\begin{lstlisting}[language=Python]
import numpy as np

def heuristics_v2(demand, capacity):
    n = demand.shape[0]
    if n == 0:
        return np.zeros((0, 0))
    
    # Algorithm parameters
    num_trials = 20
    utilization_weight = True
    perturb_factor = 0.1  # 0.0 = strict FFD, 1.0 = fully shuffled
    
    co_occurrence = np.zeros((n, n), dtype=np.float32)
    item_indices = np.arange(n)
    
    for _ in range(num_trials):
        # Create a perturbed order: mostly sorted descending, with controlled randomness
        sorted_idx = np.argsort(-demand)  # descending by size
        num_perturb = int(perturb_factor * n)
        if num_perturb > 0:
            # Randomly swap some adjacent items in the sorted list
            perm = sorted_idx.copy()
            for _ in range(num_perturb):
                i = np.random.randint(0, n - 1)
                perm[i], perm[i + 1] = perm[i + 1], perm[i]
        else:
            perm = sorted_idx
        
        shuffled_demand = demand[perm]
        shuffled_indices = item_indices[perm]
        

        bins = []  # each bin is list of original indices
        bin_loads = []  # track current load for efficiency
        
        for idx, size in zip(shuffled_indices, shuffled_demand):
            placed = False
            for b_idx, b in enumerate(bins):
                if bin_loads[b_idx] + size <= capacity:
                    b.append(idx)
                    bin_loads[b_idx] += size
                    placed = True
                    break
            if not placed:
                bins.append([idx])
                bin_loads.append(size)
        
        # Update co-occurrence with optional utilization weighting
        for b, load in zip(bins, bin_loads):
            b = np.array(b)
            weight = load / capacity if utilization_weight else 1.0
            co_occurrence[b[:, None], b] += weight
    
    # Normalize by total possible weight per trial (max weight per trial = num_trials if unweighted, or sum of bin utilizations otherwise)
    if utilization_weight:
        # Compute expected max: each trial contributes at most sum(bin_loads)/capacity = total_demand/capacity, but we normalize by trials for stability
        compatibility = co_occurrence / num_trials
    else:
        compatibility = co_occurrence / num_trials
    
    # Ensure diagonal is 1
    np.fill_diagonal(compatibility, 1.0)
    
    # Symmetrize for numerical robustness
    compatibility = (compatibility + compatibility.T) / 2.0
    
    # Row-wise normalization to [0,1]
    row_max = compatibility.max(axis=1, keepdims=True)
    epsilon = 1e-8
    normalized = np.divide(
        compatibility,
        row_max,
        out=np.zeros_like(compatibility),
        where=row_max > epsilon
    )
    
    return normalized
\end{lstlisting}

\subsection{Best-Performing Heuristic for Constructive KP}
\begin{lstlisting}[language=Python]
import numpy as np

def select_next_item_v2(remaining_capacity, values, weights):
    epsilon = 1e-9
    
    feasible_mask = weights <= remaining_capacity + epsilon
    feasible_indices = np.where(feasible_mask)[0]
    
    if len(feasible_indices) == 0:
        return 0

    n_items = len(values)
    ratios = values / (weights + epsilon)
    
    # Lookahead simulation (same as original for consistency)
    lookahead_scores = np.zeros(len(feasible_indices))
    for i, idx in enumerate(feasible_indices):
        new_cap = remaining_capacity - weights[idx]
        if new_cap < epsilon:
            lookahead_scores[i] = values[idx]
        else:
            mask = np.ones(n_items, dtype=bool)
            mask[idx] = False
            avail_w = weights[mask]
            avail_v = values[mask]
            avail_r = ratios[mask]
            
            res_feas = avail_w <= new_cap + epsilon
            if not np.any(res_feas):
                lookahead_scores[i] = values[idx]
            else:
                w_sub = avail_w[res_feas]
                v_sub = avail_v[res_feas]
                r_sub = avail_r[res_feas]
                
                order = np.argsort(-r_sub)
                total_val = values[idx]
                cap_left = new_cap
                for j in order:
                    if w_sub[j] <= cap_left + epsilon:
                        total_val += v_sub[j]
                        cap_left -= w_sub[j]
                        if cap_left < epsilon:
                            break
                lookahead_scores[i] = total_val

    # Normalize lookahead scores
    max_look = np.max(lookahead_scores)
    norm_look = lookahead_scores / (max_look + epsilon) if max_look > epsilon else np.ones_like(lookahead_scores)

    # Residual capacity with tunable exponent (soft penalization)
    residual_caps = remaining_capacity - weights[feasible_indices]
    residual_frac = np.clip(residual_caps / (remaining_capacity + epsilon), 0, 1)
    residual_exp = 0.939665
    ineff = residual_frac ** residual_exp

    # Entropy-modulated density weighting
    item_ratios = ratios[feasible_indices]
    global_max_ratio = np.max(ratios) + epsilon
    rel_density = item_ratios / global_max_ratio
    
    # Compute entropy of normalized ratios over feasible set
    prob = rel_density / (np.sum(rel_density) + epsilon)
    entropy = -np.sum(prob * np.log(prob + epsilon))
    entropy_temp = 1.733771
    entropy_mod = np.exp(-entropy / (entropy_temp + epsilon))  

    # Adaptive density factor with entropy modulation
    avg_rel_density = np.mean(rel_density) + epsilon
    base_density_factor = np.clip(rel_density / avg_rel_density, 0.6, 1.5)
    density_factor = base_density_factor * (1.0 + 0.3 * (1.0 - entropy_mod))  # boost when low entropy (more certainty)

    # DE-inspired adaptive gamma
    gamma_base = 0.276467
    choice_pressure = min(1.0, len(feasible_indices) / n_items)
    de_scale = 0.906848
    gamma = gamma_base * (1.0 + de_scale * choice_pressure)

    # Penalty term incorporating all hybrid elements
    penalty = gamma * ineff * (1.0 - norm_look) / density_factor

    # Final score: lookahead minus adaptive penalty
    scores = norm_look - penalty

    best_local = np.argmax(scores)
    return int(feasible_indices[best_local])
\end{lstlisting}

\subsection{Best-Performing Heuristic for CVRP-ACO}
\begin{lstlisting}[language=Python]
import numpy as np
import random

def heuristics_v2(distance_matrix, coordinates, demands, capacity):
    n = len(demands)
    if n <= 1:
        return np.zeros((n, n))
    
    # Hybridized parameters from Island 5 + adaptive tuning
    num_ants_factor = 3
    max_iter = 30
    alpha = 1.439913
    beta = 1.816340
    rho = 0.379887
    elite_ratio = 0.562122
    decay_factor = 1.444863
    angular_tolerance = np.pi / 4  # ~45 degrees
    
    num_ants = max(20, min(100, n * num_ants_factor))
    
    tau = np.full((n, n), 1.0 / (n * n))
    np.fill_diagonal(tau, 0)
    
    best_cost = float('inf')
    best_routes = None
    
    depot_coord = np.array(coordinates[0])
    
    # Precompute angles from depot
    angles = np.zeros(n)
    for i in range(1, n):
        vec = np.array(coordinates[i]) - depot_coord
        angles[i] = np.arctan2(vec[1], vec[0])
    
    # Adaptive neighborhood: use median instead of fixed percentile for robustness
    nonzero_dists = distance_matrix[distance_matrix > 0]
    influence_radius = np.median(nonzero_dists) if len(nonzero_dists) > 0 else 1.0
    
    # Demand density using adaptive radius
    demand_density = np.zeros(n)
    for i in range(1, n):
        neighbors = [j for j in range(1, n) if distance_matrix[i][j] <= influence_radius]
        if neighbors:
            demand_density[i] = np.mean([demands[j] for j in neighbors])
        else:
            demand_density[i] = demands[i]
    if demand_density.ptp() > 0:
        demand_density = (demand_density - demand_density.min()) / demand_density.ptp()
    else:
        demand_density = np.ones(n)
    
    # Enhanced visibility matrix
    visibility = np.zeros((n, n))
    base_demand = np.mean(demands[1:]) + 1e-8
    for i in range(n):
        for j in range(n):
            if i == j or demands[j] <= 0 or distance_matrix[i][j] == 0:
                continue
            inv_dist = 0.776078 / distance_matrix[i][j]
            urgency = demands[j] / capacity
            density_factor = 1.134148 + demand_density[j]
            
            if i == 0:
                angular_factor = 0.263295
            else:
                angle_diff = abs(angles[i] - angles[j])
                angle_diff = min(angle_diff, 2*np.pi - angle_diff)
                # Use cosine similarity within tolerance window
                if angle_diff <= angular_tolerance:
                    angular_factor = 0.405164 + np.cos(angle_diff)
                else:
                    angular_factor = np.exp(- (angle_diff / angular_tolerance)**2)
            
            visibility[i][j] = inv_dist * (1.0 + urgency) * angular_factor * density_factor

    all_route_costs_history = []
    all_routes_history = []
    
    for iteration in range(max_iter):
        all_routes = []
        all_costs = []
        
        for ant in range(num_ants):
            unvisited = set(range(1, n))
            routes = []
            total_cost = 0.0
            
            while unvisited:
                route = [0]
                load = 0
                current = 0
                
                while True:
                    feasible = [j for j in unvisited if load + demands[j] <= capacity]
                    if not feasible:
                        break
                    
                    probs = []
                    for j in feasible:
                        pher = tau[current][j] ** alpha
                        vis = visibility[current][j] ** beta
                        probs.append(pher * vis)
                    
                    if sum(probs) == 0:
                        next_node = random.choice(feasible)
                    else:
                        probs = np.array(probs) / sum(probs)
                        next_node = np.random.choice(feasible, p=probs)
                    
                    route.append(next_node)
                    load += demands[next_node]
                    unvisited.remove(next_node)
                    current = next_node
                
                route.append(0)
                routes.append(route)
                cost = sum(distance_matrix[route[i]][route[i+1]] for i in range(len(route)-1))
                total_cost += cost
            
            all_routes.append(routes)
            all_costs.append(total_cost)
            all_route_costs_history.append(total_cost)
            all_routes_history.append(routes)
            
            if total_cost < best_cost:
                best_cost = total_cost
                best_routes = routes
        
        # Pheromone evaporation
        tau *= (1 - rho)
        
        # Elite update with deduplication
        elite_count = max(1, int(elite_ratio * len(all_costs)))
        sorted_indices = np.argsort(all_costs)
        elite_indices = sorted_indices[:elite_count]
        
        seen_signatures = set()
        unique_elites = []
        for idx in elite_indices:
            sig = tuple(sorted(tuple(sorted(r)) for r in all_routes[idx]))
            if sig not in seen_signatures:
                seen_signatures.add(sig)
                unique_elites.append(idx)
        
        # Temporal-decay visitation entropy
        visit_count = np.zeros(n)
        total_visits = 0
        recent_solutions = all_routes_history[-min(20, len(all_routes_history)):]
        for t, routes in enumerate(reversed(recent_solutions)):
            weight = decay_factor ** t
            for route in routes:
                for node in route[1:-1]:
                    visit_count[node] += weight
                    total_visits += weight
        
        if total_visits > 0:
            visit_prob = visit_count / total_visits
            visit_prob = np.clip(visit_prob, 1e-10, None)
            visit_prob /= visit_prob.sum()
            elite_entropy = -np.sum(visit_prob * np.log(visit_prob + 1e-10))
        else:
            visit_prob = np.ones(n) / n
            elite_entropy = np.log(n)
        
        # Pheromone update with entropy scaling
        for rank, idx in enumerate(unique_elites):
            routes = all_routes[idx]
            cost = all_costs[idx]
            if cost == 0:
                continue
            delta_tau = (1.0 + elite_entropy) / (cost * (rank + 1))
            for route in routes:
                for i in range(len(route) - 1):
                    u, v = route[i], route[i+1]
                    tau[u][v] += delta_tau
                    tau[v][u] += delta_tau
        tau = (tau + tau.T) / 2

    edge_scores = np.zeros((n, n))
    
    if best_routes is None:
        return edge_scores

    # Final entropy for scoring
    visit_count = np.zeros(n)
    total_visits = 0
    recent_solutions = all_routes_history[-min(20, len(all_routes_history)):]
    for t, routes in enumerate(reversed(recent_solutions)):
        weight = decay_factor ** t
        for route in routes:
            for node in route[1:-1]:
                visit_count[node] += weight
                total_visits += weight
    
    if total_visits > 0:
        visit_prob = visit_count / total_visits
        visit_prob = np.clip(visit_prob, 1e-10, None)
        visit_prob /= visit_prob.sum()
        entropy = -np.sum(visit_prob * np.log(visit_prob + 1e-10))
    else:
        visit_prob = np.ones(n) / n
        entropy = np.log(n)
    
    # Compute centroids and slack
    route_centroids = []
    route_slack = []
    for r in best_routes:
        nodes = [node for node in r if node != 0]
        if nodes:
            coords = np.array([coordinates[node] for node in nodes])
            centroid = np.mean(coords, axis=0)
            load = sum(demands[node] for node in nodes)
        else:
            centroid = depot_coord
            load = 0
        route_centroids.append(centroid)
        route_slack.append(capacity - load)
    
    best_total_cost = sum(
        sum(distance_matrix[r[i]][r[i+1]] for i in range(len(r)-1))
        for r in best_routes
    )
    
    base_weight = (1.0 / (best_total_cost + 1e-8)) * (1.0 + entropy)
    
    # Score edges
    for idx, route in enumerate(best_routes):
        centroid = route_centroids[idx]
        slack = route_slack[idx]
        route_nodes = [node for node in route if node != 0]
        
        if route_nodes:
            node_coords = np.array([coordinates[node] for node in route_nodes])
            std_dev = np.std(np.linalg.norm(node_coords - centroid, axis=1)) if len(route_nodes) > 1 else 1.0
        else:
            std_dev = 1.175239
        
        for i in range(len(route) - 1):
            u, v = route[i], route[i+1]
            dist = distance_matrix[u][v]
            if dist == 0:
                continue
            
            # Radial alignment with cosine similarity
            if u != 0 and v != 0:
                vec_u = np.array(coordinates[u]) - depot_coord
                vec_v = np.array(coordinates[v]) - depot_coord
                norm_u = np.linalg.norm(vec_u)
                norm_v = np.linalg.norm(vec_v)
                if norm_u > 0 and norm_v > 0:
                    cos_sim = np.dot(vec_u, vec_v) / (norm_u * norm_v)
                    radial_score = 0.406854 + max(0.0, cos_sim)
                else:
                    radial_score = 0.984288
            else:
                radial_score = 0.300000
            
            # Compactness based on centroid proximity
            mid_point = (np.array(coordinates[u]) + np.array(coordinates[v])) / 2.0
            compactness = np.exp(-np.linalg.norm(mid_point - centroid) / (std_dev + 1e-6))
            
            # Slack utilization factor
            slack_factor = 1.352515 + (slack / capacity)
            
            # Distance penalty
            dist_penalty = 2.423188 / (dist + 1e-8)
            
            # Frequency from visit probability
            freq_factor = (visit_prob[u] + visit_prob[v]) / 2.0
            
            weight = base_weight * radial_score * compactness * slack_factor * dist_penalty * freq_factor
            edge_scores[u][v] += weight
            edge_scores[v][u] += weight
    
    total_score = edge_scores.sum()
    if total_score == 0:
        return edge_scores
    return edge_scores / total_score
\end{lstlisting}

\subsection{Best-Performing Heuristic for Car Mountain}
\begin{lstlisting}[language=Python]
import numpy as np
import random
import math
def choose_action_v2(pos, v, last_action):
    # KEY PARAMETERS
    goal_threshold = 0.5
    min_velocity_base = 0.013
    climb_vel_boost = 0.025
    left_zone_base = -0.45
    coast_zone = 0.05
    left_push_sensitivity = 0.8
    stop_tolerance_factor = 6

    # Normalize position to [0, 1] for dynamic adjustments
    norm_pos = (pos + 1.2) / 1.8
    dynamic_min_velocity = min_velocity_base + climb_vel_boost * norm_pos

    # Near goal: stabilize or gently accelerate right
    if pos >= goal_threshold:
        return 1 if abs(v) < dynamic_min_velocity else 2

    # If moving right fast enough, keep accelerating
    if v > dynamic_min_velocity:
        return 2

    # If moving left significantly, accelerate left to deepen swing
    if v < -dynamic_min_velocity / stop_tolerance_factor:
        return 0

    # Adjust left push boundary based on current velocity (slower = push earlier)
    effective_left_boundary = left_zone_base + left_push_sensitivity * abs(v)

    # Decision based on adjusted zones and velocity state
    if pos < effective_left_boundary:
        # Deep in left: start right swing
        return 2
    elif pos < 0.0:
        # On left slope: push left to gain potential energy
        return 0
    elif pos < coast_zone and abs(v) < dynamic_min_velocity:
        # Near bottom with low speed: coast to preserve momentum
        return 1
    else:
        # On right slope or with decent speed: accelerate right
        return 2
\end{lstlisting}

\subsection{Best-Performing Heuristic for DPP-GA}
\begin{lstlisting}[language=Python]
import numpy as np

def crossover_v2(parents, n_pop, blend_prob=0.5, crossover_points=2):
    """
    Generate offspring via hybrid crossover: arithmetic blend or multi-point segment crossover.
    
    Parameters:
        parents (np.ndarray): 2D array of shape (n_parents, n_decap)
        n_pop (int): number of offspring to produce
        blend_prob (float): probability of using arithmetic blending
        crossover_points (int): number of crossover points for segment crossover
    
    Returns:
        np.ndarray: offspring array of shape (n_pop, n_decap)
    """
    n_parents, n_decap = parents.shape
    offspring = np.empty((n_pop, n_decap), dtype=parents.dtype)

    # Pre-generate all random choices for efficiency
    rand_vals = np.random.rand(n_pop)
    parent_indices = np.random.choice(n_parents, size=(n_pop, 2), replace=True)
    
    # Ensure distinct parents per offspring
    equal_mask = parent_indices[:, 0] == parent_indices[:, 1]
    while np.any(equal_mask):
        parent_indices[equal_mask, 1] = np.random.randint(0, n_parents, size=np.sum(equal_mask))
        equal_mask = parent_indices[:, 0] == parent_indices[:, 1]

    p1 = parents[parent_indices[:, 0]]
    p2 = parents[parent_indices[:, 1]]

    # Arithmetic blending mask
    blend_mask = rand_vals < blend_prob

    # Apply arithmetic blend where applicable
    if np.any(blend_mask):
        alphas = np.random.rand(np.sum(blend_mask), 1)
        offspring[blend_mask] = alphas * p1[blend_mask] + (1 - alphas) * p2[blend_mask]

    # Segment crossover for the rest
    segment_mask = ~blend_mask
    if np.any(segment_mask):
        num_segment = np.sum(segment_mask)
        seg_p1 = p1[segment_mask]
        seg_p2 = p2[segment_mask]
        child_seg = np.empty((num_segment, n_decap), dtype=parents.dtype)

        for i in range(num_segment):
            # Generate unique sorted crossover points
            if n_decap > 1 and crossover_points > 0:
                pts = np.sort(np.random.choice(n_decap - 1, size=min(crossover_points, n_decap - 1), replace=False)) + 1
            else:
                pts = np.array([], dtype=int)
            segments = np.split(np.arange(n_decap), pts)
            
            use_p1 = True
            for seg in segments:
                if use_p1:
                    child_seg[i, seg] = seg_p1[i, seg]
                else:
                    child_seg[i, seg] = seg_p2[i, seg]
                use_p1 = not use_p1
        offspring[segment_mask] = child_seg

    return offspring
\end{lstlisting}







\end{document}